\documentclass[a4paper,12pt]{amsart}
\input epsf
\usepackage{epsfig}
\usepackage{hyperref}
\usepackage{caption}
\usepackage{subcaption}
\usepackage{float}
\usepackage{caption}
\usepackage[ruled]{algorithm2e}
\usepackage{mathtools}
\usepackage{tikz}
\DeclareMathAlphabet\mathbold{OML}{cmm}{b}{it}
\numberwithin{equation}{section}

\newtheorem{theorem}{Theorem}[section]

\newtheorem{definition}{Definition}[section]

\def\b0{{\mathbf 0}}
\def\b1{{\mathbf 1}}

\newcommand{\R}{{\mathcal R}}
\newcommand{\A}{{\mathcal A}}
\newcommand{\B}{{\mathcal B}}
\newcommand{\C}{{\mathcal C}}
\newcommand{\D}{{\mathcal D}}

\newcommand{\I}{{\mathcal I}}

\newcommand{\M}{{\mathcal M}}

\renewcommand{\L}{{\mathcal L}}
\renewcommand{\P}{{\mathcal P}}
\newcommand{\W}{{\mathcal W}}

\begin{document}

\title{Extending bootstrap AMG for clustering of\\ attributed graphs}

%% use optional labels to link authors explicitly to addresses:
%% \author[label1,label2]{}
%% \affiliation[label1]{organization={},
%%             addressline={},
%%             city={},
%%             postcode={},
%%             state={},
%%             country={}}
%%
%% \affiliation[label2]{organization={},
%%             addressline={},
%%             city={},
%%             postcode={},
%%             state={},
%%             country={}}

%\author{}

%\affiliation{organization={},%Department and Organization
%            addressline={},
%            city={},
%            postcode={},
%            state={},
%            country={}}

\author[P.~D'Ambra]{Pasqua D'Ambra}
\author[P.~S.~Vassilevski]{Panayot S. Vassilevski}
\author[L.~Cutillo]{Luisa Cutillo}

\address[P.~D'Ambra]{pasqua.dambra@cnr.it, Institute for Applied Computing (IAC), CNR, Via P. Castellino, 111, 80131, Napoli, Italy}
\address[P.~S.~Vassilevski]{panayot@pdx.edu, Fariborz Maseeh Department of Mathematics and Statistics, Portland State University, Portland, OR 97207, USA}
\address[L.~Cutillo]{l.cutillo@leeds.ac.uk, School of Mathematics, University of Leeds, Leeds, LS2 9JT, UK}

\begin{abstract}
In this paper we propose a new approach to detect clusters in undirected graphs with attributed vertices. We incorporate structural and attribute similarities between the vertices in an augmented graph by creating additional vertices and edges as proposed in~\cite{CZY2011,ZCY2009}.  The augmented graph is then embedded in a Euclidean space associated to its Laplacian and we cluster vertices via a modified K-means algorithm, using a new vector-valued distance in the embedding space. Main novelty of our method, which can be classified as an {\em early fusion method}, i.e., a method in which additional information on vertices are fused to the structure information before applying clustering, is the interpretation of attributes as new realizations of graph vertices, which can be dealt with as coordinate vectors in a related Euclidean space. This allows us to extend a scalable generalized spectral clustering procedure which substitutes graph Laplacian eigenvectors with some vectors, named {\em algebraically smooth vectors}, obtained by a linear-time complexity Algebraic MultiGrid (AMG) method.

We discuss the performance of our proposed clustering method by comparison with recent literature approaches and public available results. Extensive experiments on different types of synthetic datasets and real-world attributed graphs show that our new algorithm, embedding attributes information in the clustering, outperforms structure-only-based methods, when the attributed network has an ambiguous structure. Furthermore, our new method largely outperforms the method which originally proposed the graph augmentation, showing that our embedding strategy and vector-valued distance are very effective in taking advantages from the augmented-graph representation.
\end{abstract}

%%Graphical abstract
%\begin{graphicalabstract}
%\includegraphics{grabs}
%\end{graphicalabstract}

%%Research highlights
%\begin{highlights}
%\item A new generalized spectral clustering method to detect community in attributed graphs.
%\item A linear complexity (scalable) algorithm, based on a bootstrap Algebraic MultiGrid method, to estimate near-kernel components of graph Laplacian needed for graph embedding in Euclidean space.
%\item A new vector-valued distance and an extended version of the classical K-means by using the new distance.
%\item Extensive experiments on different types of synthetic datasets and real-world attributed graph datasets widely used in the literature demonstrates improvement with respect to the state of the art in detecting clusters in graphs with ambiguous structure.
%\item A freely-available software framework to be used in clustering of graphs also with node attributes.
%\end{highlights}


\maketitle

\section{Introduction}
\label{intro}

Graph clustering is a main task in network analysis~\cite{S2007,F2010,NC2011}. In many real-world networks, such as social and biomedical networks, in addition to the graph structure some information related to possible vertex features are often available; such additional information, often referred to as metadata, can be used to improve the accuracy of clustering algorithms. Many clustering methods have been extended to incorporate features information in the vertex similarities detection so that the clusters are not only densely connected but also very homogeneous in their attributes, as discussed in section~\ref{relwork}.
This paper describes an overall approach that enables clustering of networks with categorical vertex attributes. Our method builds on a recently proposed generalized spectral method~\cite{DCV2019}. In the following we introduce the general settings and briefly summarize our approach.

Given a set of vertices $V$ and a set of $m$ categorical labels $L=\{ l_1, \ldots, l_m \}$, a vertex-attributed graph is the 4-tuple $G=(V,E,\W,L)$, where $E \subset V \times V$ is the edge set and $\W=(w_{ij})_{(i,j) \in E}$ is the (positive) edge weights matrix. Each vertex $v_i \in V$ is associated with an attribute vector $(l_1(v_i), \ldots, l_m(v_i))$, where $l_j(v_i)$ is the attribute value of vertex $v_i$ on attribute $l_j$. The cardinality or size of $G$ is the total number of vertices $|V|=n$.

In~\cite{DCV2019} we formulated the clustering of an undirected graph, as a partition $V_1, \ldots, V_K$, in $K$ nonempty sets, so that $V_k \subset V$, $V_k \cap V_h= \emptyset, \ \forall k \neq h$ and $V_1 \cup \ldots \cup V_K=V$, which minimizes the following functional:
\begin{equation}
RatioCut(V_1, \ldots, V_K)= \frac{1}{2} \sum_{k=1}^K \frac{W(V_k, \overline{V_k})}{|V_k|},
\label{ratiocut}
\end{equation}
where $W(V_k, \overline{V_k})= \sum_{i \in V_k, j \in \overline{V_k}} w_{ij}$, and $\overline{V_k}$ is the complement of $V_k$ in $V$. We observe that a graph partition which minimizes~(\ref{ratiocut}) corresponds to a partition which minimizes the weight of the edges between two different sets and maximizes the number of the vertices within a set.
The solution of (a relaxed version of) this problem involves the eigenvectors of the (weighted) graph Laplacian that are used to define embedding of the graph into a Euclidean space of low dimension and use geometric means to construct clusters~\cite{HTF2001}.
In~\cite{DCV2019}, we have generalized this concept by looking at the spectrum of a generalized eigenvalue problem involving the (weighted) graph Laplacian and an adaptive Algebraic MultiGrid (AMG) operator (linear solver) which provided a natural way to select the dimension of the embedding Euclidean space. Our main motivation was to design a linear complexity algorithm which can be more scalable than standard spectral clustering approaches which require computation of graph Laplacian eigenvectors. The method demonstrated a very good ability to detect clusters in graphs with medium and high modularities, i.e., when a block structure is well defined.

In this work, we extend the method to incorporate additional information on graph vertices, so that the ability to detect clusters in graphs with a low modularity can be improved while preserving the linear complexity.
More precisely, the aim is to re-formulate the method, to obtain partitions where the vertices within one set have similar attribute values. In order to achieve this purpose, we introduce an approach that starts with the augmentation of the graph $G$ with new vertices representing the attribute values and new edges connecting original vertices to new vertices representing their attribute values. After this augmentation, we solve the minimization problem associated to the relaxed form of the functional~(\ref{ratiocut}) (as in~\cite{DCV2019}) for the augmented graph, this creates a low dimensional embedding space associated with the Laplacian of the augmented graph. A modified K-means algorithm is then applied in the embedding space, where a new distance measure between two original vertices is defined.

%\subsection*{MAIN CONTRIBUTIONS SUMMARY}
The main original contribution and results of this paper can be summarised as follows:
\begin{itemize}
\item We propose a new generalized spectral clustering method to detect community in  attributed graphs; it belongs to the class of {\em early fusion methods}, where attributes are fused to the structure information before applying the embedding in a Euclidean space associated to some generalized eigenvectors of the Laplacian graph. The new method allows us to extend the embedding procedure introduced in~\cite{DCV2019}, showing flexibility and robustness of the procedure itself.
\item We introduce a new vector-valued distance in the embedding space, which has its rationale into a block representation of the added (attribute) vertices with their structure vertex origin, since they represent a particular feature of their origin; then we extended the classical K-means by using the new distance.
\item Extensive experiments on different types of synthetic datasets and real-world attributed graph datasets widely used in the literature show that our new algorithm employing attributes information improves the ability of the structure-only-based methods, when the attributed network has an ambiguous structure, with significant advantage with respect to state-of-the-art structure-based methods. Comparisons with results available in the literature also demonstrate that our new method largely outperforms the method which originally proposed the graph augmentation employed in this paper, showing that our embedding strategy and vector-valued distance are very effective in taking advantages from the augmented-graph representation.
\item We finally made available all the functionalities used in our algorithm in a freely-available software framework to be used in clustering of attributed graphs.
\end{itemize}

The paper is organized as follows. In Section~\ref{relwork} we present related work on various different approaches to fuse structure and attribute and classify our method with respect to well recognized current taxonomy. In Section~\ref{laplacianembedding} we summarize main features of a clustering algorithm already proposed by us to partition un-directed graphs and whose extension to attributed graphs is the objective of this work. In Section~\ref{augmentedgraphs} we provide a description of a strategy which augments the original graph with new vertices and new edges depending on the attribute values associated to the original graph vertices.

%*******remove

%It was previously proposed in~\cite{ZCY2009,CZY2011}, and can also be viewed as a {\em coarse} version of a graph disaggregation (cf.,~\cite{KV2013}).
%The authors in~\cite{ZCY2009,CZY2011} defined a distance measure based on random walks on the augmented graph which allowed them to then apply a geometric (spatial) clustering, i.e., K-means thus extending the concept of closeness to vertices sharing the same label value. In Section~\ref{sub:newdistance}, starting from a representation of original graph vertices in terms of a set of coordinates which is obtained by embedding the augmented graph in a vector space associated to its Laplacian, and then includes information on attributes, we define an extension of the usual Euclidean distance, so that a version of K-means can be applied to identify partitions of the attributed graph having homogeneous labels.

%**********

In Section~\ref{BCM4G} we describe the architecture of the proposed software framework implementing all the functionalities needed to apply the method and analyse its asymptotic computational complexity. In Section~\ref{res} we present the experimental setting and discuss results on synthetic graphs and on some real-world data sets. Section~\ref{conc} contains concluding remarks and possible future works.

\section{Related Work}
\label{relwork}

The growing need for embedding vertex attributes in graph clustering generates a large amount of research methods aiming to fuse information on graph structure and vertex features.
Some interesting review are in~\cite{BCMM2015,C2020}. An important issue is related to the correlation between vertex attributes and graph structure. It is a general belief that the presence of structure-attributes correlation improves the quality of the clustering, however some recent works seem to refuse this assumption at least for some type of fusion methods~\cite{CGB2020}. This discussion is beyond the scope of our work, and we focus on analyzing the state-of-the-art techniques able to add vertex attributes in the cluster detection.
In~\cite{BCMM2015} the authors classify the proposed fusion methods into
different categories depending on the adopted clustering approach, and indeed many authors, driven by the clustering method, agree in classifying the methods in two main classes named {\em distance-based} and {\em model-based}~\cite{BAGC2012,BAGC2014,HuangYLLC17}. The aim of distance-based methods is to define a form of unified distance which combine structural and attribute information, and then leverage on existing clustering methods, such as K-means, functional optimization, spectral clustering, to cluster attributed graphs based on the unified distance. The unified distance can be defined in an implicit way, e.g., by introducing weights on nodes~\cite{Neville03}, or explicitly, e.g., by distance functions~\cite{Combe2012,ZCY2009,CZY2011}. Model-based methods mainly rely on statistical inference and setup a generative model which well fits both the structural and vertex attributes information~\cite{BAGC2012,BAGC2014,NC2016}.

In~\cite{C2020} we can find an interesting prior classification depending on when structure and attributes are fused with respect to the clustering algorithm. Although the author focuses on some aspects of social networks, his
review is very general and exhaustive. Three overall classes of methods are considered: {\em early fusion methods}, if structure information and attributes are fused before applying a clustering algorithm,
{\em simultaneous fusion methods} that fuse structure and attributes simultaneously while applying the clustering algorithm and finally, {\em late fusion methods} that apply clustering to the graph structure and to the attribute labels separately and finally fuse the obtained groups. The definition of these three classes is useful
to understand the level of influence of the attribute information on the method, and it clarifies that
methods belonging to the early fusion and late fusion classes have often the advantage to reuse approaches that already demonstrated their effectiveness in clustering of graphs with no use of additional information on vertices.
The methods are also classified in terms of the methodology applied to find clusters. For the detailed classification we refer the reader to~\cite{C2020}, which also includes tables of main methods belonging to each class, and an interesting ranking of the most influential methods in each class, based on their PageRank values.
On the base of the above taxonomy, our approach belongs to the class of early fusion methods, and going down to the technique-related subclasses, we can classify it in the subclass of {\em node-augmented graph-based}. Main methods belonging to this class are the {\em SA-cluster}~\cite{ZCY2009,CZY2011} and its improved versions. We will give details on the above methods in section ~\ref{augmentedgraphs}. We also observe that, our approach actually combines different methodologies intersection of three subclasses, including also the sub-classes of embedding-based and distance-based methods, as it is described in the following sections.
%Nella classe dei simultaneous fusion ci sono quelli basati su approccio statistico bayesiano (BCAG), quelli basati su ottimizzazione (I-Louvain) anche con approcci euristici-evoluzionari, approach based on matrix compression (see PICS)

\section{Embedding in the space of algebraically smooth vectors for graph Laplacian}
\label{laplacianembedding}

In~\cite{DCV2019} we proposed a new method for efficient embedding of undirected graphs in a low-dimensional Euclidean space spanned by an accurate approximation of the smallest part of the spectrum of a linear operator associated to the graph Laplacian. The method can be viewed as an extension of the classical spectral clustering which employs a standard eigenvalue problem for graph Laplacian.

Let $\D=diag(d_i)_{i=1}^n$, with $d_i= \sum_{j=0}^n w_{ij}$, be the diagonal matrix of weighted vertex degrees of $G$, the graph Laplacian of $G$ is the matrix $\L=\D-\W \in \R^{n \times n}$. We observe that $\L$ is a symmetric, positive semi-definite M-matrix, and its spectrum is a main tool for studying the graph~\cite{vL2007}. In particular, in spectral clustering, a graph is embedded in the Euclidean space spanned by the first $K$ eigenvectors corresponding to the first $K$ smallest eigenvalues of $\L$; this embedding allows to apply standard spatial clustering algorithms, e.g., the K-means optimization algorithm, giving good approximations of optimal partitioning for well-clustered graphs~\cite{PSZ2015}.

As low-dimensional space for graph embedding, we use the space spanned by some vectors, referred to as {\em algebraically smooth vectors} generated by an iteration process. We use the bootstrap process coming from an Algebraic MultiGrid (AMG) operator associated to the graph Laplacian. Note that we always consider the graph Laplacian of a connected graph; in the case of graphs with more than one connected component, we apply our method to each of its connected components. Furthermore, we also eliminate singularity of the graph Laplacian by a rank-1 update which gives symmetric positive definite (s.p.d.) graph Laplacian matrix $\L_{S}=\L+\lambda \; e \cdot e^T$, where $e$ is a vector of dimension $n$ having unit (nonzero) components only for a single pair of indices $i$ and $j$ corresponding to an arbitrarily chosen edge $(i,j) \in E$. In the following, we briefly summarize the theoretical basis of our method and its formulation.

In the theory of AMG~\cite{MLBFP2008}, an $\epsilon-${\em smooth vector} of the matrix $\L_S$ is defined as follows:
\begin{definition}\label{def:algebraic_smooth_vectors}
	Let $\M$ be a s.p.d. smoothing operator, i.e., the iteration matrix of a relaxation method convergent in the $\L_S$-norm\footnote{The $\A-$norm of the vector $u$, where $\A$ is a s.p.d. matrix is defined as $\|u\|_{\A}= \sqrt{u^T \A u}$.}, and given $\epsilon \in (0,1)$, we say that the vector $u$ is algebraically $\epsilon$-smooth with respect to $\L_S$~if
	\begin{equation*}
	\|(I- \M^{-1}\L_S) u\|^2_{\L_S} \ge (1-\epsilon) \|u\|^2_{\L_S}.
	\end{equation*}
	We note that $\|(I- \M^{-1}\L_S) u\|_{\L_S}\le \|u\|_{\L_S}$ (since $\M$ is a convergent smoother).
	That is, $u$ is in the near-nullspace of $\M^{-1}\L_S$ in the sense that $\M^{-1}\L_S u$ has a much smaller norm than $u$.
\end{definition}
Such a vector can be obtained by performing a few iterations of the smoother $\M$, i.e.
\begin{equation}
u=u^t=(I-\M^{-1}\L_S)u^{t-1},
\label{smooth vector based on M}
\end{equation}
starting from a random vector $u^0$ for a sufficiently large $t$.

The above concept was generalized in~\cite{DCV2019} in the sense that we used an operator $\B$ (in fact a sequence of operators that gradually approximate $\L_S$) in place of $\M$ and considered the respective algebraically smooth vectors.

Given the matrix $\L_S$, we first define a two-level method for solving the linear system $\L_S x = b$, where $x \in \R^n$ is the unknown vector and $b \in \R^n$ is a given right-hand side term.
Let $V_c \subset \R^{nc}$ be the space spanned by a set of smooth vectors $\{u_r\}_r$ of $\L_S$ with respect to the operator $\M$. Let $\{\phi_i\}_{i=1}^{n_c}$ provide a basis of $V_c$, and let $\P = [\phi_1,\;\dots,\; \phi_{n_c}]$ be the interpolation matrix mapping vectors from $V_c$ to $V \subset \R^{n}$ and $(\L_S)_c = \P^T\L_S\P$ the corresponding coarse matrix. We define a two-level operator $\B$ corresponding to the iteration matrix
\begin{equation*}
    I - \B^{-1}\L_S = (I-\M^{-T}\L_S)(I-\P (\L_S)^{-1}_c \P^T\L_S) (I-\M^{-1}\L_S),
\end{equation*}
which leads to the formula
\begin{equation*}
\B^{-1} = \overline{\M}^{-1} + (I-\M^{-T}\L_S) \P (\L_S)^{-1}_c \P^T (\I-\L_S\M^{-1}),
\end{equation*}
where $\overline{\M} = \M (\M+\M^T-\L_S)^{-1}\M^T$ is the so-called symmetrized smoother, such that $\I-\overline{\M}^{-1}\L_S = (\I-\M^{-T}\L_S)(\I-\M^{-1}\L_S)$.

From the general theory of AMG methods~\cite{MLBFP2008} we can infer the following spectral equivalence result for the operator $\B$:
\begin{equation*}
v^T \L_S v \le v^T \B v \le C\; v^T \L_S v, \ \ \forall v \in V
\end{equation*}
where the optimal constant $C$ is given by the formula:
\begin{equation}\label{Cconstant}
C = \max\limits_{v \in V}\;\frac{\|v-\pi v\|^2_\M}{\|v\|^2_{\L_S}}.
\end{equation}
Here $\pi:V \;  \mapsto V_c \subset V$ is a projection in the $\M-$inner product, i.e., $\pi=\P(\P^T\M\P)^{-1}\P^T\M$.
Furthermore, the following theorem is proved in~\cite{DCV2019}:
\begin{theorem}\label{spectralconstant}
Consider the operator $\B$ defined by the two-level algorithm with smoother $\M$, symmetrized smoother $\overline{\M}$, and coarse space $V_c$. Let $\rho$ be the convergence rate of the iterative process having $\B$ as iteration matrix. Consider the generalized eigenvalue problem:
\begin{equation}\label{geigenp}
\L_S q_k = \lambda_k \M q_k,
\end{equation}
where $0 \le \lambda_1 \le \lambda_2 \le \dots \le \lambda_s < \lambda_{s+1} \le \dots \le \lambda_{\max}$,
and assume that the coarse space $V_c$ contains the first $s \ge 1$ eigenvectors $q_k$.
Then, the following estimate holds for the spectral equivalence constant $C$:
\begin{equation*}
C \le \frac{1}{1 -\rho} \simeq \frac{1}{\lambda_{s+1}}.
\end{equation*}
\end{theorem}

The above result assures that if $\B$ is a method with a sufficiently small convergence rate, then eigenvectors in~(\ref{geigenp})
can effectively be replaced (as argued in~\cite{DCV2019}) by the algebraically smooth vectors generated by the iteration process based on $\B$.
More specifically, in~\cite{DCV2019} we proposed an iterative procedure which justified the use, in place of the eigenvectors $q_k$, of a sequence of smooth vectors with respect to $\L_S$. These vectors are obtained during a bootstrap procedure whose final aim was to construct a composite AMG operator $\B$ leading to a  sufficiently small spectrally equivalence constant $C$.
For sake of completeness, we summarize here main features of the bootstrap procedure and the resulting graph embedding algorithm.

The bootstrap procedure builds a composite linear solver $\overline{\B}$ obtained from a multiplicative composition of a number of AMG operators $\B_r$. The operator $\overline{\B}$  is defined from the error propagation matrix:
\begin{equation}
\label{comp_AMG}
I - \overline{\B}^{-1} \L_S = (I-\B^{-1}_0 \L_S)(I-\B^{-1}_1 \L_S)\dots\; (I-\B^{-1}_r \L_S),
\end{equation}
where each $\B_r$ is an AMG operator, e.g., a standard V-cycle involving a simple smoother such as the weighted Jacobi or the Gauss-Seidel method, and a hierarchy of coarse matrices exploiting  {\em coarsening based on compatible weighted matching} in the form studied in~\cite{DV2013,DFV2018,DV2019}.
To construct the operator $\B_r$, we first generate an algebraically smooth vector $u_r$. The first smooth vector is constructed based on the smoother $\M$ for $\L_S$ as described in~\eqref{smooth vector based on M}. At step $r>0$, we apply few iterations based on the currently available composite solver ${\overline \B}$ consisting of the already built $r-1$ components; i.e., $u=u^t=(I-\overline{\B}^{-1}\L_S)u^{t-1}$, starting from a random vector $u^0$. Once the vector $u_r$ is available, the operator $\B_r$ is defined as an AMG cycle where the construction of its hierarchy of coarse spaces is driven by the vector $u_r$ using {\em coarsening based on compatible weighted matching}~\cite{DV2013,DFV2018,DV2019}.

The process is iterated several times until $M$ vectors $\{u_r\}_{r=1}^M$ are computed; the number $M$ of computed vectors depends on the desired convergence factor $\varrho$ that we impose on the composite solver $\B_{BAMG}=\overline{\B}$ in~(\ref{comp_AMG}) to exhibit  after $M$ composite steps.
As a result, for a given $\varrho \in (0,1)$, the number $M$ of components (equivalently the number of smooth vectors associated with the components $\B_r$) are such that by construction, we have:
\begin{equation*}
\label{BAMG}
\|\I-\B^{-1}_{BAMG}\L_S\|_{\L_S} \le \varrho.
\end{equation*}
In order to generate a set of orthogonal vectors spanning the graph embedding space, we finally apply a Singular Value Decomposition (SVD) to the computed set of vectors $\{u_r \}^M_{r=1}$ and use the corresponding left-singular vectors $\{\overline{u}_r\}_{r=1}^{n_c}$ as basis for the embedding space.
In addition to this set we add the vector of all ones, which is the null vector of $\L$, and we also use this vector as the starting one in the bootstrap process.
After embedding, we apply any standard spatial clustering procedure to identify possible vertex clusters of the original graph.
Our method of choice is the widely used K-means algorithm, as detailed in~\cite{DCV2019} (see Algorithm 2).

Note that our procedure exploits increasingly accurate
coarse spaces which provide improved approximations to the generalized eigenvalue problem~(\ref{geigenp}). This is the case since the operator $\B$ changes at each bootstrap step until we reach a final operator which has a sufficiently small spectral equivalence constant~(\ref{Cconstant}). The achieved convergence tolerance determines how many bootstrap steps to perform, and as a result gives a criterion how to choose the dimension of the embedding space. For the more standard spectral clustering it is not as clear how many eigenvectors to compute in advance in order to get satisfactory clustering results. Therefore, our method exhibits a desired property in practice and we view it as an advantage over the more standard spectral clustering approaches.
%We additionally observe that our method has a linear computational complexity with a proportionality constant that grows with the number of built smooth vectors which makes our approach more competitive for large networks.

In the following, we describe the extension of the method based on the bootstrap AMG to undirected graphs with vertex attributes.

\section{Attribute-augmented Graphs}
\label{augmentedgraphs}

Let $G = (V, E, \W, L)$ be a vertex-attributed graph, where $L$ is the set of attribute values, i.e., the set of vectors $(l_1(v_i), \ldots, l_m(v_i))$ with $m$ values associated to each vertex $v_i \in V$, and let $Dom(l_j) = \{ l_{j,1}, \ldots, l_{j,n_j} \}$ be the $n_j$ distinct values of the attribute $l_j$.
Algorithm \ref{alg:aug}
builds an augmented graph with additional vertices $V_a$, edges $E_a$ and edge weights $W_a$. It implements the method from ~\cite{ZCY2009,CZY2011}, and can be viewed also as a coarsened version of an attribute-based graph disaggregation algorithm~\cite{KV2013}.
\begin{algorithm}[htb]
\SetAlgoLined
\KwData{vertex-attributed graph $G=(V,E,\W,L)$, $n=|V|$, $ne=|E|$, $m$: number of label classes in $L$, $weight(\cdot)$: real function depending on vertices and attribute values}
\KwResult{augmented graph $G_A^H=(V \cup V_a, E \cup E_a, \W_a)$, $n_{new}=|V \cup V_a|$, $ne_{new}=|E \cup E_a|$}
 \For{$j=1, \ldots, m$}
 {
   identify $Dom(l_j)=\{l_{j,1}, \ldots, l_{j,n_j} \}$ as the set of distinct values of the label $l_j$\;
}
$E_a=\emptyset$, $V_a=\emptyset$, $\W_a=\W$\;
\For{$j=1, \ldots, m$}
{
\For{$k=1, \ldots, n_j$}
{
$V_a=V_a \cup \{ v_{j,k} \}$\;
\For{$i=1, \ldots, n$}
{
\If{$l_j(v_i)==l_{j,k}$}
{
$E_a= E_a \cup \{e_{new}=(v_i, v_{j,k}) \}$\;
$\W_a(e_{new})=weight(e_{new})$\;
}
}
}
}
$n_{new}=n+\sum_{j=1}^m n_j$\; $ne_{new}=ne+nm$\;
\caption{\small{Graph augmentation based on vertex attributes.}\label{alg:aug}}
\end{algorithm}
In summary, Algorithm~\ref{alg:aug} builds the augmented graph defined as follows:
\begin{definition}\label{auggraphzhou}
Let $G = (V, E, \W, L)$ be a vertex-attributed graph, where $L$ is the set of attribute values, i.e., the set of vectors $(l_1(v_i), \ldots, l_m(v_i))$ with $m$ values associated to each vertex $v_i \in V$, and let $Dom(l_j) = \{ l_{j,1}, \ldots, l_{j,n_j} \}$ be the $n_j$ distinct values of the attribute $l_j$.
 The augmented graph
$G_A^H=(V \cup V_a, E \cup E_a, \W_a)$, constructed in Algorithm~\ref{alg:aug} has its components defined as:
\begin{enumerate}
\item $V_a=(v_{j,k})$ is the set of vertices representing the distinct attribute values $(l_{j,k})_{j=1,k=1}^{m, n_j}$ of the vertices in $G$;
\item $E_a$ is the set including the edges $(v_{i},v_{j,k}) \ \forall j=1, \ldots, m$ iff $l_j(v_i)=v_{j,k}$, which connect the vertex $v_i$ with his attribute value $v_{j,k}$;
\item $\W_a$ is a real matrix including possible weights for the edges in $E \cup E_a$.
\end{enumerate}
\end{definition}
We note that the cardinality of the augmented graph $G^H_A$ is equal to $n_{new}=n+\sum_{j=1}^m n_j$, where $n_j \leq n$ is the number of different values of the attribute $l_j$, and the number of edges is $ne_{new}=ne+nm$, where $ne$ is the number of edges in $G$. We also observe that in our implementation of the method and in the experiments discussed in section~\ref{res} we consider unweighted graphs, where matrix $\W$ and $\W_a$ corresponds to the adjacency matrix of the original graph and of the augmented graph, respectively. On the other hand, the extension of the code to the weighted graphs is straightforward, while the impact of possible weights on the edges could have large impact on the final clusterings for any employed method.
\begin{figure}[htb]
\centering
\includegraphics[width=0.49\textwidth]{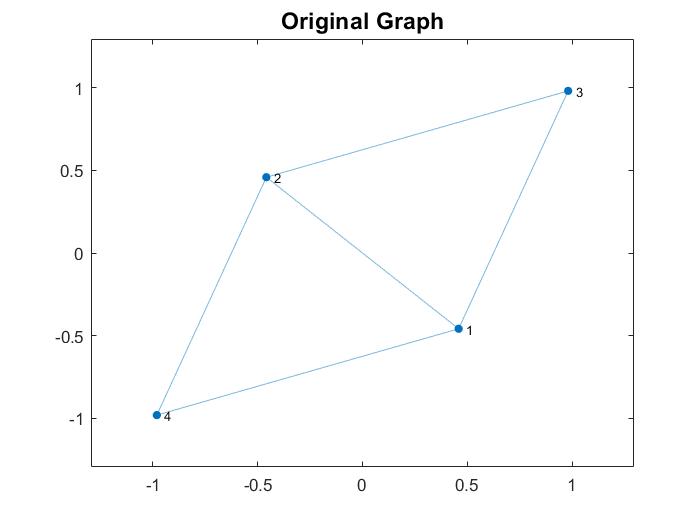}
\includegraphics[width=0.49\textwidth]{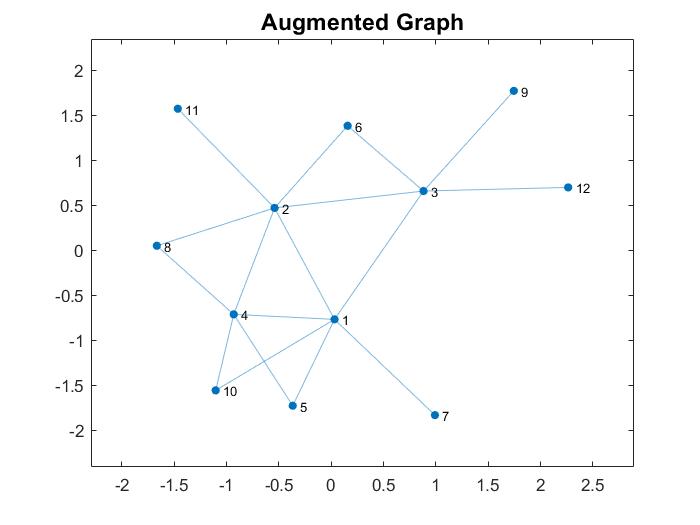}
\caption{\small{Original (left) and Augmented Graph (right).}\label{toygraph}}
\end{figure}
In Fig.~\ref{toygraph} we show a small undirected graph with $3$ different labels and the corresponding augmented graph. The original graph has the vertex set $V=\{v_1=1,v_2=2,v_3=3,v_4=4\}$ and the edge set $E=\{(1,2), (1,3), (1,4), (2,4), (2,3)\}$, then $|V|=n=4$ and $|E|=ne=5$; the vertices are characterized by the labels defined in Table \ref{tab:toylabels}.
\begin{table}[h]
    \centering
    \begin{tabular}{c|ccc}
    \hline
    vertex & \multicolumn{3}{c}{labels} \\
    \hline
     & $l_1$ & $l_2$& $l_3$ \\
     \cline{2-4}
    1      & M & R & C \\
    2      & F & D & P \\
    3      & F & I & J \\
    4      & M & D & C\\
    \hline
    \end{tabular}
    \caption{\small{Sample Graph: Vertex Attributes}}
    \label{tab:toylabels}
\end{table}
In this case, we have $m=3$, $Dom(l_1)=\{M,F\}$, $Dom(l_2)=\{ R,D,I\}$, $Dom(l_3)=\{ C, P, J\}$, $n_1=2$ and $n_2=n_3=3$. Algorithm \ref{alg:aug} builds the set of $8$ new vertices, enumerated with consecutive integers from $5$ to $12$. We have $V_a=\{v_{1,1}=M=5, v_{1,2}=F=6, v_{2,1}=R=7, v_{2,2}=D=8, v_{2,3}=I=9, v_{3,1}=C=10, v_{3,2}=P=11, v_{3,3}=J=12\}$. The set of new edges reads: $E_a= \{(1,5), (1,7), (1,10), (2,6), (2,8), (2,11),(3,6), (3,9), (3,12), (4,5), (4,8), (4,10)\}$. Finally, the augmented graph $G^H_A$ has a total of $n_{new}=12$ vertices and a total of $ne_{new}=17$ edges. Both, the original graph and the augmented graphs have no edge weights.

In the augmented graph (see Definition~\ref{auggraphzhou}) two original vertices $v_{i_1}$ and $v_{i_2}$, referred to as {\em structure vertices}, can be close either if they are connected through other structure vertices, or if they share attribute vertices as neighbors, or both. In~\cite{ZCY2009,CZY2011} the authors have defined a unified distance measure on the augmented graph using random walks, which estimates the pairwise vertex closeness in the graph through both structure and attribute edges. Using their distance measure, they apply a modified K-means algorithm constituting the so-called Structural/Attribute clustering algorithm ({\em SA-cluster}).

In this paper, we follow a different approach which can be seen as a generalization of a spectral clustering algorithm extending the method proposed in~\cite{DCV2019} to deal with vertex-attributed graphs. In details, after augmentations we compute the s.p.d modified graph Laplacian $\L_s$ of the augmented graphs $G^H_A$ and embed the graphs in the subspace spanned by the smooth vectors of $\L_S$ associated to a composite AMG operator $B_{BAMG}$ as in~(\ref{BAMG}). These smooth vectors, having $n_{new}$ entries are used to represent the original $n$ structure vertices in terms of a set of $1+m$ scalar components, where the first component is related to the original graph structure information and the resulting $m$ are related to the vertex attributes.
In this way, each original (structure) vertex is associated to $m$ additional (attribute) vertices, and can be viewed as a node of $m+1$ disaggregated vertices (in analogy to~\cite{KV2013}).
Another analogy can be seen with the representation of a vector field in a physical (2D or 3D) space domain where the value of the field in a space point is represented by a node vector of (2 or 3) scalar components.
The latter point of view was our motivation to define in section~\ref{sub:newdistance} the distance metric~\eqref{distblock}, which is another key ingredient of our method.

Before focusing on our approach to identify partitions in the structure vertices after embedding of augmented graph, we point out that there is also another potential feature in exploiting the augmented graph with the goal to analyze important properties of the original vertex-attributed graph, that is the detection of overlapped clusters. To illustrate the idea, consider our example in Fig.~\ref{toygraph}.
%\begin{itemize}
 %   \item
 We can use any existing clustering technique applied to the augmented graph, e.g., the method we studied in~\cite{DCV2019} or any other available method. %popular Louvain algorithm \cite{Blondel2008}.
 Assume that one of these methods partitioned the augmented graph into the following two clusters:
    \begin{equation*}
        \A^a_1 = \{1,4,5,7,8,10\} \text { and } \A^a_2= \{2,3,6,9,11,12\}.
    \end{equation*}
    If we restrict these sets to the original (structure) vertices, we get the following two clusters:
    \begin{equation*}
        \A_1 = \{1,4\} \text { and } \A_2= \{2,3\},
    \end{equation*}
    which provides one possible way to construct clusters for vertex-attributed graphs. In the present paper, we cluster the original graph by embedding the augmented one into an Euclidean space of some dimension and then define distance between two structure (original) vertices only, so that a new distance-based K-means is applied to them. We observe that this approach is computationally less expensive than applying classical K-means to all vertices of the augmented graph. However, assume that we did cluster the augmented graph leading to the clusters $\A^a_1$ and $\A^a_2$. It allows us to additionally exploit the attributed vertices as explained in the following.
    %Note that the new vertices came from some structure vertices.
    If we replace the new (attribute) vertices with their related structure vertices in the sets $\A^a_1$ and $\A^a_2$, i.e., $5 \mapsto 1,4$; $6 \mapsto 2, 3$; $7 \mapsto 1$; $8 \mapsto 2,4$; $9 \mapsto 3$; $10 \mapsto 1,4$; $11 \mapsto 2$, and $12 \mapsto 3$, we end up with the following new partitioning of the original structure vertices:
    \begin{equation*}
        {\widetilde \A}_1 = \{1,2,4\} \text { and } {\widetilde \A}_2= \{2,3,4\}.
    \end{equation*}
    We notice that we identified two overlapping clusters. Therefore, the augmentation process, if fully utilized, offers a potential to create overlapping subgroups of structure (original) vertices.

\subsection{Vertex block-coordinates and new vector-valued Euclidean distance measure}
\label{sub:newdistance}

%In the case of our first augmentation approach, we consider the augmented graph $G_A$ associated to the vertex attributed graph $G$, let $w_r  \in \mathcal{R}^{n(q+1)}$ be a smooth vector w.r.t. the augmented graph laplacian of $G_A$ obtained as explained in sec.~\ref{laplacianembedding}.
%We can form the coordinate vectors $v^r_{i}= \left . w_r \right |_{\tau_i}=(w^r_{i,j})_{j=1}^{q+1}$ per each super-vertex $v_{i}$ and per each smooth vector $w_r$.

\subsubsection{Vertex block-coordinates} \label{subsection: vertex blockcoordinates}
Starting from the augmented graph $G_A^H$ associated with the attributed graph $G$, let %$\overline{u}_r=(\overline{u}^r_1, \ldots, \overline{u}^r_n, \ldots, \overline{u}^r_{n_{new}}) \in \R^{n_{new}}$
$\overline{u}_r=(\overline{u}^r_1, \ldots, \overline{u}^r_n, \ldots, \overline{u}^r_{n_{new}}) \in \R^{n_{new}}$ be a smooth vector computed for the augmented graph Laplacian of $G_A^H$. We recall that the vertices of the augmented graph come with double subscripts; namely,
structure vertex $v_i$ is indexed as
$v_{i,1}$ whereas all its attribute neighbors $v_{j,k}$ are indexed such that for attribute $j=1,2,...,m$ we have $l_j(v_i) = v_{j,k}$ (the value from the domain of $l_j$). According to this convention (applying it to the indices of the smooth vectors), we can partition the values of any smooth vector in blocks, where each block will correspond to a structure vertex and its $m$ attribute neighbors. That is, we can form block coordinate vectors $v^r_{i}=(\overline{u}^r_{i,j})_{j=1}^{m+1}$ for each structure vertex $v_{i}$ and each smooth vector $\overline{u}_r$ where, as already indicated, we let  $\overline{u}^r_{i,1}=\overline{u}^r_i$, and $\overline{u}^r_{i,j+1}=\overline{u}^r_{j,k}$ if $l_j(v_i)=v_{j,k}$, $\forall j=1, \ldots, m$. In this way, the original graph vertices will have the same coordinate $\overline{u}^r_{j,k}$ if they share the same attribute value $v_{j,k}$ on a given attribute $l_j$, so that vertices sharing an attribute value belong to the same hyperplane corresponding to that value.

\begin{figure}[htb]
\centering
\includegraphics[width=0.75\textwidth]{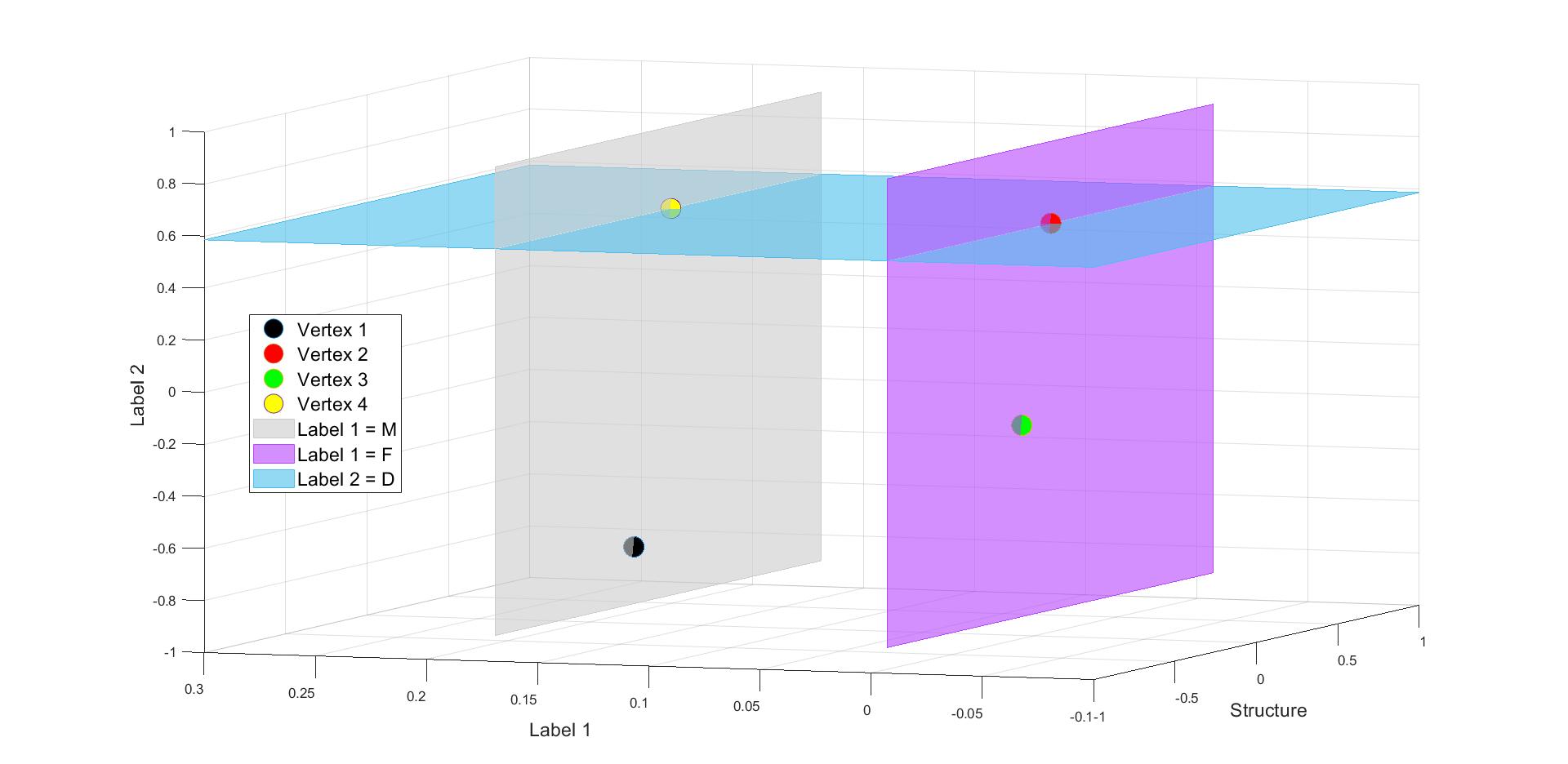}
\includegraphics[width=0.75\textwidth]{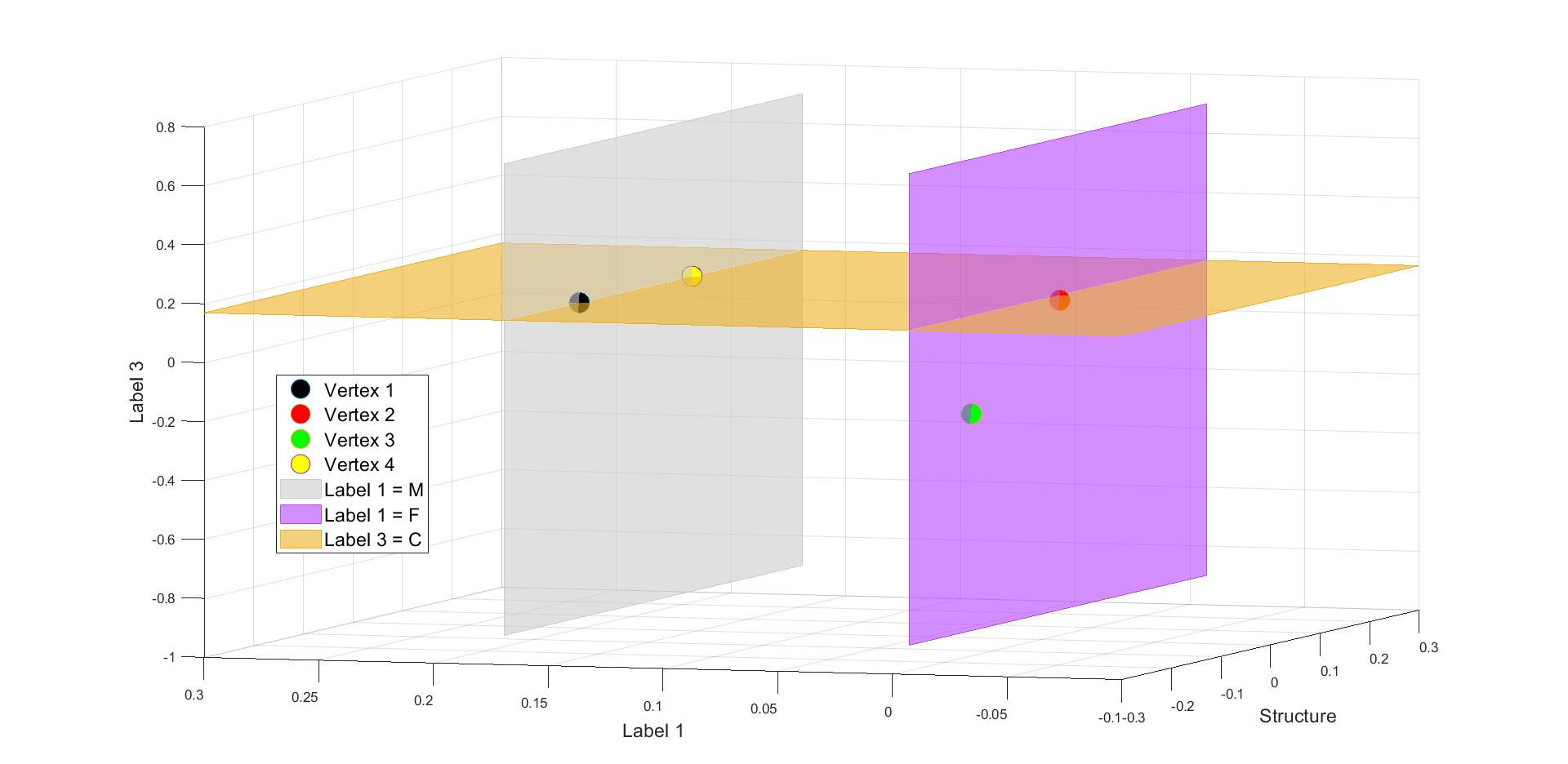}
\includegraphics[width=0.75\textwidth]{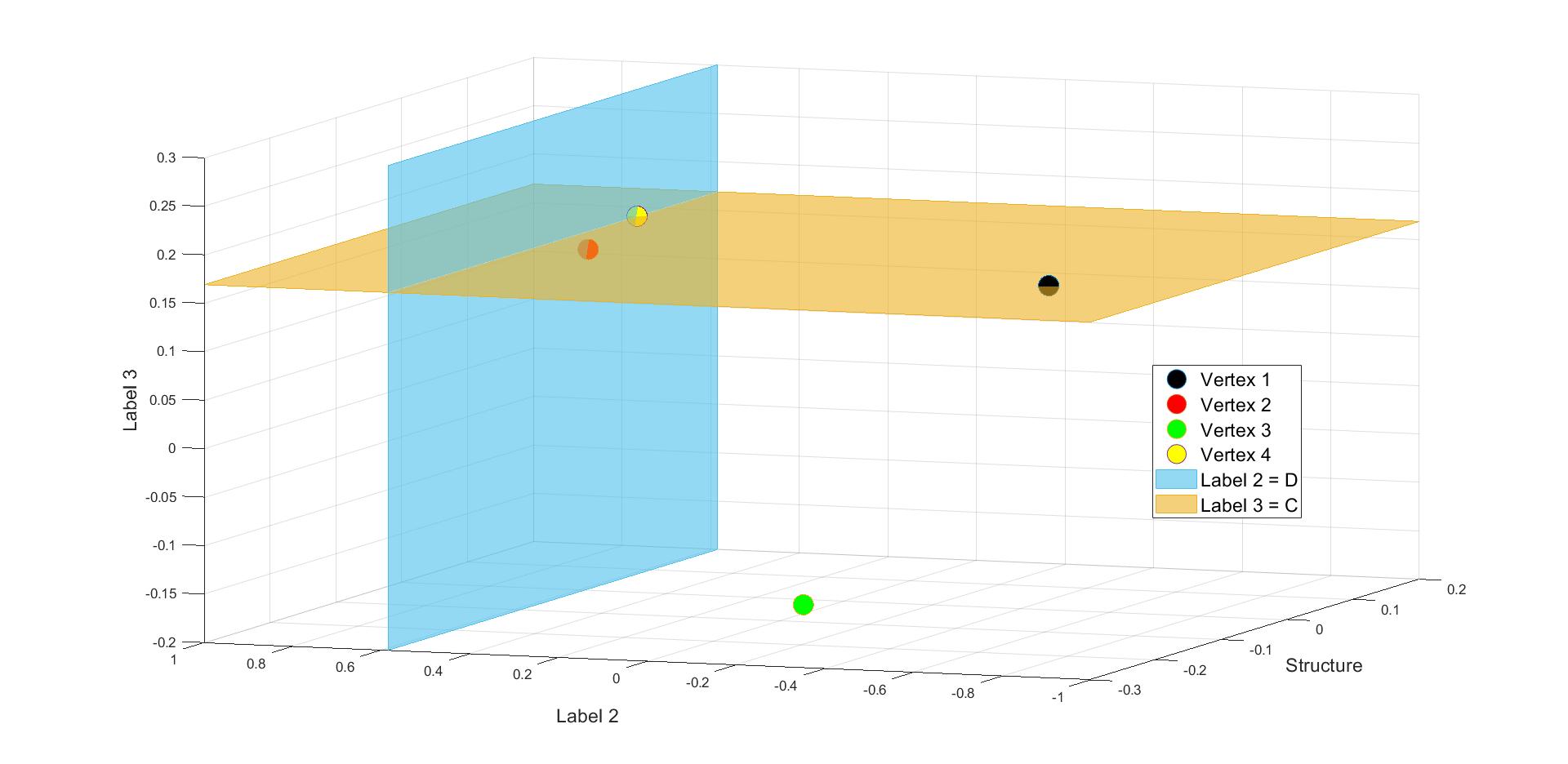}
\caption{\small{Coordinates representation for the sample graph: the $4$ coordinates of each original graph vertex, corresponding to $1$ smooth vector, are represented in 3D, fixing on $x-$axis the coordinate related to the structure information and varying on $y-$axis and $z-axis$ the coordinates related to different labels. Planes corresponding to the same attribute values are also drawn.}\label{toygraph:coordinates}}
\end{figure}

In Fig.~\ref{toygraph:coordinates} we draw in a 3D space a picture of the coordinates representing the $4$ original graph vertices obtained by using only $1$ smooth vector generated by the bootstrap algorithm applied to the augmented graph Laplacian. We fix the coordinate related to the structure information on $x$-axis and use a combination of the other $3$ coordinates associated to the attributes as $y-$ and $z-$coordinates. We also draw the planes corresponding to the same attribute values, so that vertices sharing attribute coordinates belong to the same plane.

\subsubsection{Distance metric}
\label{subsection: distance metric}
After assigning to each structure vertex $v_i$ a block  ``augmented'' smooth vector component $v^r_i \in {\R}^{1+m}$, it is straightforward to define a distance between any two structure vertices $v_{i_1}$ and $v_{i_2}$ based on these $(1+m)-$vector coordinates.

Let $n_c$ be the number of smooth vectors obtained by the bootstrap AMG applied to the augmented graph Laplacian $\L_S$ and let $v_{i_1} = (\overline{u}^r_{i_1,j})^{m+1,n_c}_{j=1,r=1}$ and $v_{i_2} = (\overline{u}^r_{i_2,j})^{m+1,n_c}_{j=1,r=1}$ be the block-coordinates of the vertices, then we let
\begin{equation}
dist(v_{i_1},v_{i_2}) = \sqrt{ \sum\limits^{n_c}_{r=1} \|\overline{u}^r_{i_1}-\overline{u}^r_{i_2}\|_2^2},
\label{distblock}
\end{equation}
where $\|\overline{u}^r_{i_1}-\overline{u}^r_{i_2}\|_2^2$ is the square of the usual Euclidean distance between the block components $(\overline{u}^r_{i_1,j})^{m+1}_{j=1}$ and  $(\overline{u}^r_{i_2,j})^{m+1}_{j=1}$.

The function in~(\ref{distblock}) is clearly a metric (due to the standard Cauchy-Schwarz inequality for vectors).

Note that if we did not have labels, i.e., if $m=0$, the above expression reduces to the standard definition of distance once we have embedded the graph in a Euclidean space. For $m>0$, our motivation is to block the added vertices with their structure vertex origin since they represent a particular feature of their origin.

%%% {\bf The function in \Ref{distblock} is a metric, indeed it is straightforward to %%% show that it verify all the properties of a metric as a simple generalization of %%% the distance based on $\ell_2-$norm:
%%% \begin{itemize}
 %%%    \item $dist(v_{i_1},v_{i_2}) \ge 0$;
 %%%    \item $dist(v_{i_1},v_{i_2}) = 0$ iff $v_{i_1}=v_{i_2}$;
    %indeed $dist(v_{i_1},v_{i_2})=0 \leftrigtharrow  %%% %%%\|\overline{u}^r_{i_1}-\overline{u}^r_{i_2}\|_2^2=0 \forall r \leftrigtharrow %%% %%%\overline{u}^r_{i_1}=\overline{u}^r_{i_1}$, for the properties of  $ell_2-$norm;
    %%% %%% \item $dist(v_{i_1},v_{i_2}) \leq dist(v_{i_1},v_{i_3}) +  %%% %%% dist(v_{i_2},v_{i_3})$
%%% %%%\end{itemize}
%%% %%% Do we have to prove or can we say that is straightforward?}
Once we have a distance measure defined, we are able to apply a K-means clustering algorithm for the set of structure vertices.

\section{BCMatch4Graphs Software Framework}
\label{BCM4G}

We implemented all the functionalities needed to apply our clustering algorithms, both for simple graphs and for vertex attributed ones, extending the software framework described in~\cite{DFV2018}. The extended code to deal with graphs has been made available with the new name \verb|BCMatch4Graphs| at~\href{https://github.com/bootcmatch/BCMatch4Graphs}{https://github.com/bootcmatch/BCMatch4Graphs}.
\begin{figure}[htb]
\centering
\includegraphics[width=0.7\textwidth]{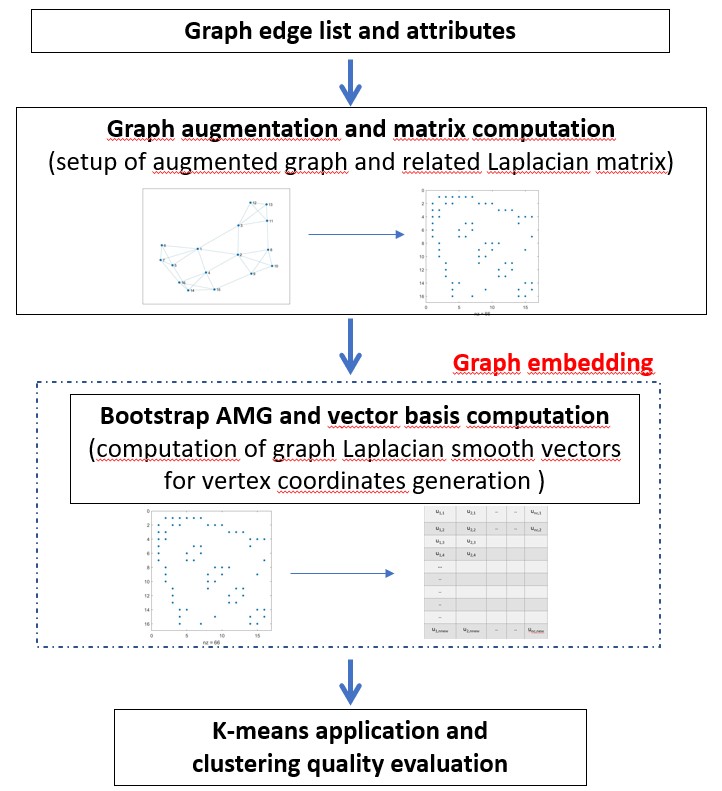}
\caption{\small{Flowchart of a typical computation workflow based on BCMatch4Graphs.}\label{fig:bcm4graphs}}
\end{figure}
In Fig.~\ref{fig:bcm4graphs} we show a flowchart of a typical computation workflow supported by the software framework whose architecture has a suitable layered structure.
We extended the basic layer of the modular BootCmatch framework with some functionalities for manipulation of un-directed graphs, computation of related matrix representations, including Laplacian matrix and rank-1 update to eliminate constant vector from its kernel. In this layer we also included the code for graph augmentation to be applied if vertex attributes are available, as described in Algorithm~\ref{alg:aug}. We note that a support function to extract from each class of attributes only their different values, i.e., the set $Dom(l_j)$, per each class label $l_j$, is included in the augmentation algorithm. All the functionalities for graph embedding are exposed from a middle software layer implementing the setup and the application of the operator in~(\ref{BAMG}), to which correspond a number of algebraically smooth vectors spanning the graph embedding space. After smooth vectors computation, a function to compute the SVD is applied to define a vector basis. Finally, functions needed to define vertex coordinates and then compute the usual Euclidean distance for simple graphs and the new vector-valued distance as defined in~(\ref{distblock}) for vertex attributed graphs, are also included in the middle layer. At the application layer, functionalities to apply
the K-means iterative procedure and evaluate clustering quality, i.e., compute clustering modularity and conditional entropy with respect to the set of attributes, as defined in~(\ref{mod}) and~(\ref{condentr}), respectively, are implemented.
Some user's level drivers to cluster un-directed connected graph, with and without vertex attributes, are also available in the framework.

\subsection{Computational complexity}
\label{sub:complexity}

We observe that the way we build and apply each component of the bootstrap AMG operator in~(\ref{BAMG}) defines the computational complexity of the graph embedding procedure and then its
convergence properties, i.e., the number of components needed to reach a desired convergence rate. The embedding software layer in \verb|BCMatch4Graphs| supports different AMG cycles among
the most widely used, featured by different computational complexities also
depending on the algorithm applied to generate each single AMG operator. Furthermore, symmetrized versions of the multiplicative composition of the built AMG components can be applied, producing a more accurate composite AMG.
The choice of accuracy of each AMG operator and the way we compose the different built operators influences the number of bootstrap steps, and then the number of computed smooth vectors needed to reach the desired convergence rate decreases. This automatic way to compute the dimension of the embedding vector space is one of the most important advantages of our approach, and the large flexibility in selecting the algorithmic parameters to setup and apply the composite operator $\overline{\B}$ defines the final computational complexity of the overall clustering procedure, which can be tuned depending on the user's needs.
Some best practice choices, as detailed in section~\ref{res}, are available so that the final embedding procedure has a linear computational complexity $\mathcal{O}(M \cdot ne_{new})$, where $ne_{new}$ is the number of edges in the augmented graph and $M$ is the number of bootstrap steps (and then of the computed smooth vectors) needed to reach the desired convergence rate. Therefore, the final computational complexity of the clustering method is $\mathcal{O}(M \cdot ne_{new}+ M \cdot n_{new} \cdot nit)$, where $\mathcal{O}(M \cdot n_{new} \cdot nit)$
is the complexity of the K-means procedure if $nit$ cluster iterations are performed.

\section{Experimental Results}
\label{res}

In the present section, we study  the clustering quality of our method described in section~\ref{augmentedgraphs} on two types of synthetic graphs widely used as benchmarks and on some real-life attributed graphs available from public repositories.
We also compare our algorithm with some existing clustering methods, whose implementation is freely available, some of which also use information on vertices attributes.
%We consider both methods that do not incorporate attribute information and others that do employ such attribute information.
More specifically, we consider the following attribute-free methods:
\begin{itemize}
\item Louvain method ($Lou$) is a greedy optimization method that attempts to maximize the modularity of a partion of the graph by a multilevel algorithm~\cite{Blondel2008};
\item Fast Greedy method ($FG$) is also a greedy method which tries to maximize modularity by a hierarchical agglomeration algorithm~\cite{CNM2004};
\item $BCMG$ is our original method, available in \verb|BCMatch4Graphs| and based on the bootstrap AMG for the Laplacian graph~\cite{DCV2019}. We point out that some default algorithmic parameters to setup the method are applied, as discussed in~\cite{DCV2019} and defined in the same sample of configuration files available at the software repository. In order to reduce sensitivity of the K-means algorithm w.r.t. the starting cluster centers, we run the k-means for $100$ times with the same input data and choose the partition corresponding to the maximum achieved value of the computed modularity.
In the application of the bootstrap AMG, we built $m$ smooth vectors, where $m= \min (40, \overline{K} )$, with $\overline{K}$ the number of
AMG operators built to obtain a final multiplicative operator of type~(\ref{comp_AMG}) with a desired convergence rate less than $\varrho=10^{-8}$. Each AMG operator is a symmetric V-cycle with one forward/backward Gauss-Seidel sweep as pre/post-smoothing. A direct LU factorization is used as coarsest solver.
\end{itemize}
Then we consider the following clustering  methods that employ information on vertices attributes:
\begin{itemize}
    \item $Newman$ is the method described in~\cite{NC2016}. It uses a techniques of Bayesian statistical inference to fit a suitable network model to the available data, trying to find a possible partition correlated to the attributes;
    \item $BAGC$ is a model-based approach to attributed graph clustering. It also relies on a Bayesian probabilistic model which tries to fit both structural and attribute aspects of networks~\cite{BAGC2012};
    \item $BCMAG$ is the new clustering method based on the augmented graph described in this paper and implemented in \verb|BCMatch4Graphs|. The bootstrap algorithm for augmented graph embedding and the modified K-means are applied in the same conditions of the above $BCMG$ method.
\end{itemize}

Note that $Lou$ and $FG$ are implemented in the $R$ package $igraph$. $BAGC$ is available on the GitHub repository~\cite{bacgcode}. Software implementing $Newman$ is available as supplementary material of~\cite{NC2016}. We observe that this software does not allow the use of more than one class of attributes. For this reason we could not show any results from $Newman$ method in Section~\ref{sub:LFR} and in some of the test cases in Section~\ref{sub:realgraphs}.

\subsection{Evaluation Metrics}
\label{sub:metrics}

In order to analyze the obtained results, we employ standard metrics which allow us to compare two different clusterings and then quantify the performance of the clustering algorithms when the ground truth is given. Let $\C'$ be the ground truth, and $\C$ be the estimated partition obtained by a given clustering method, in the following we define {\em Normalized Mutual Information} (introduced in~\cite{Dannon2005}), {\em entropy} and {\em gain}.

Given the graph $G$ and two partitions $\C ={C_1,\ldots,C_K}$ and $\C'={C'_1,\ldots,C'_{K'}}$, with $K$ and $K'$ non empty sets, respectively, we consider
\begin{itemize}
    \item $p(k)$ to be the probability of a vertex to belong to group $C_k$.
    \item $p(k,k')$ to be the probability that a vertex belongs to set $C_k$ from $\C$ and to the group $C'_{k'}$ from partition $\C'$.
\end{itemize}
If we set $n_k=|\C_k|$ and $n_{k,k'}=|\C_k \cap \C_{k'}|$, we can define $p(k)=n_k/n$ and  $p(k,k')= n_{k,k'}/n$.

$NMI$ is defined as:
\[
NMI(\C,\C')= \frac{I(\C,\C')}{H(\C) + H(\C')},
\]
where $H(C)$ is the entropy associated with partition $\C$:
\begin{equation}\label{entr}
H(\C)=-\sum_{k=1}^K p(k) \log(p(k)),
\end{equation}
and $I(C,C')$ is the mutual information between $\C$ and $\C'$, i.e., the information that one partition has about the other:
\[
I(\C,\C')= \sum_{k=1}^K \sum_{k'=1}^{K'} p(k,k') \log \frac{p(k,k')}{p(k) p(k')}.
\]
Note that $NMI \in [0,1]$, and two partitions can be considered in good agreement when $NMI \approx 1$.

Given the entropy $H(\C)$ and $H(\C')$ of two clusterings, the cluster-specific entropy of $\C'$ , that is the conditional entropy of $\C'$ with respect to cluster $C_k$, is defined as:
\[
H(\C'|C_k)=-\sum_{k'=1}^{K'} \frac{n_{k,k'}}{n_k} \log\left(\frac{n_{k,k'}}{n_k}\right),
\]
%.....https://www.dcs.bbk.ac.uk/~ale/dsta/2020-21/dsta-8/zaki-meira-ch1cerpt.pf
The {\em conditional entropy} of $\C'$  given clustering $\C$ is then defined as the weighted sum:
\begin{equation}
\label{condentr}
H(\C'|\C)=\sum_{k=1}^K \frac{n_k}{n}H(\C'|C_k)=-\sum_{k=1}^K\frac{n_k}{n}\sum_{k'=1}^{K'} \frac{n_{k,k'}}{n_k} \log\left(\frac{n_{k,k'}}{n_k}\right).
\end{equation}

Expression~(\ref{condentr}) suggests that the more cluster's members are split into different partitions, the higher the conditional entropy. If $\C'$ represents the ground truth, for a perfect clustering the conditional entropy value, which for the sake of brevity we refer to as entropy in the rest of the paper, should be zero.\\ %It can be shown also that:
%\begin{equation*}
%    H(\C'|\C)= H(\C,\C')-H(C),
%\end{equation*}
%where $ H(\C,\C')=-\sum_{k=1}^K\sum_{k'=1}^{K'} P(k,k') \log  P(k,k')$ is the joint entropy of $\C'$ and $\C$.It follows that The conditionalentropy $H(\C'|\C)$  measures the remaining entropy of $\C'$ given $\C$.
%In particular, $H(\C'|\C) = H(\C)$ if $\C$ provides no information about $\C'$ .
   %https://rstudio-pubs-static.s3.amazonaws.com/455435_30729e265f7a4d049400d03a18e218db.html

Suppose we want to measure if a certain partition reduces the overall entropy and hence is more informative. The measure that we suggest to look at is called {\em information gain}, or more simply, {\em gain}. Information gain is the expected reduction in entropy caused by partitioning the vertices according to a
given partition $\C$. It is defined as the difference between the absolute entropy of the clustering $\C'$ and the conditional entropy  $H(\C'|\C)$:
\[
IG(\C',\C)=H(\C')-H(\C'|\C)
\]

\subsection{SBM Graphs}
\label{sub:SBM}

In order to assess the overall proposed methodology we started from a Stochastic Block Model (SBM)~\cite{holland1983} to generate random graphs. This type of synthetic graphs are often used to identify specific  model parameters that are critical for the ability of an algorithm to find reliable partitions.

Assume there are $q$ sets of vertices in a graph, a SBM is specified by the expected fraction of vertices in each set $n_a, ~ 1\le a \le q$, and by the probability $p_{ab}$ of an edge between set $a$ and set $b$ $ \forall a,b \in [1,\ldots,q]$. These probabilities form a matrix $P$ that is commonly referred to as affinity matrix.  We focus on a simple well-known (cf.,~\cite{Decelle2011}) SBM model called planted graph.
The planted model is a special SBM case in which the values of the probability in the affinity matrix $P$ are equal to a constant $p_{in}$ on the diagonal and equal to another constant $p_{out}$ off the diagonal; thus, two vertices within the same set share an edge with probability $p_{in}$, while two vertices in different sets share an edge with probability $p_{out}$. In particular, if $p_{in} > p_{out}$ the model is called assortative, while if $p_{in}<p_{out}$ the model is called dis-assortative.
In our simulations we assume an assortative structure, so that $p_{in} > p_{out}$.
We generate a SBM random graph $G$ of $n$ vertices partitioned into $q$ sets, $C_1,\ldots,C_q$, with adjacency matrix $\A_{ij} = 1$ if there is an edge between vertices $i$ and $j$, and $0$ otherwise; more  specifically, we consider Bernoulli trials for each potential edge with the probabilities given by the affinity matrix, so that the corresponding adjacency matrix has entries defined as follows:
\[
\A_{rs}|~ r \in C_a \: s \in C_b \sim Bern(p_{ab}), ~\forall r,s \in [1,\ldots,n]
\]
where for $a,b \in [1,\ldots, q]$:
\[
 p_{ab}  =
    \begin{cases}
      p_{in} & \text{if  $a=b$ }\\
      p_{out}  & \text{otherwise}\\
    \end{cases}
\]
Note that the probabilities are defined in such a way that the average vertex degree is a given $c$. In particular,  the parameter \textbf{c} can be used to modulate the sparsity of the simulated networks.
Each vertex has a label in $[1,\ldots,q]$, indicating which set it belongs to.
A study presented in~\cite{Decelle2011} identifies conditions under which polynomial-time algorithms can find a partition that is correlated with the planted partition. More specifically this is possible when
    \begin{equation}\label{eq:thresh1}
p_{in}-p_{out} >\frac{\sqrt{q p_{in} + q(q - 1)p_{out}}}{\sqrt{n}}.
\end{equation}

This implies that, depending on the generative parameter values, this condition might not be satisfied. It is often convenient to work with a scaled affinity matrix so that $p_{in}=\frac{c_{in}}{n}$ and $p_{out}=\frac{c_{out}}{n}$, where $c_{in}$ is the average vertex degree of vertices having all edges within the same set and $c_{out}$ is the average vertex degree of vertices having edges between two different sets.
In this case, the average degree of the graph can be expressed as:
\begin{equation}
c=\frac{q-1}{q}c_{out}+\frac{1}{q}c_{in}.
\label{eq:c}
\end{equation}
Rewriting (~\ref{eq:thresh1}) in terms of $c_{in}$ and $c_{out}$, we have:
    \begin{equation}\label{eq:cinout}
c_{in}-c_{out}>\sqrt{q[c_{in} + (q-1) c_{out}]},
\end{equation}
and from (\ref{eq:c}) we finally obtain:
    \begin{equation}
c_{in}-c_{out}>q\sqrt{c}.
\end{equation}\label{eq:dt}

We observe that when $c_{in}$ is much greater than $c_{out}$ the graph shows a strong cluster structure, while such structure becomes weaker when $c_{in}$ is close to $c_{out}$. If follows that we can implicitly tune the strength of the set/cluster structure by varying the difference $c_{in}-c_{out}$.

The strength of division of a graph into clusters can be measured by the {\em modularity}~\cite{Newman2006}. It can be broadly defined as the fraction of the edges that fall within the given set minus the expected value of such fraction if edges were distributed at random, in details:
\[
    Q=\frac{1}{2ne}\sum_{ij}[\A_{ij}-\frac{c_ic_j}{2ne}]\delta(C_i,C_j),
    \label{mod}
\]
where $ne$ is the number of edges in the whole graph, $c_i$, $c_j$ are the degree of vertices $i$ and $j$ respectively,  $\A=(\A_{ij})_{ij}$ is the adjacency matrix of the graph, $\delta(C_i, C_j)$ is equal to $1$ if vertices $i$ and $j$ belong to the same cluster and $0$ otherwise.
Graphs with high modularity have dense connections between vertices within clusters but sparse connections across clusters. We observe that $Q \in [-1,1]$ and it is often used to evaluate the quality of a partition obtained by a clustering method. Furthermore, many clustering methods are based on optimization algorithms which try to maximize modularity, e.g. {\em Lou} and {\em FG}.

In some cases it is expected that a graph has natural partition into clusters referred to as ground truth. Clustering methods aim to estimate such a {\em true} partition. In our simulation framework the ground truth is known and a high modularity corresponds to $c_{in}>>c_{out}$, while a low modularity corresponds to $c_{in} \sim c_{out}$.
We varied the difference $c_{in}-c_{out}$ above and below the so-called {\em detectability threshold} $q\sqrt{c}$, given a specific set of average degrees $c\in [5,10,15,20]$ and number of clusters $q \in [2,3,4,5]$. In the following discussion, we use the true labels, i.e., the index of the cluster that a vertex belongs to as vertex attribute, so vertices in the same cluster have the same attribute.

\begin{figure*}[htb]
        \centering
        \begin{subfigure}[b]{0.485\textwidth}
            \centering
            \includegraphics[page=1, width=\textwidth]{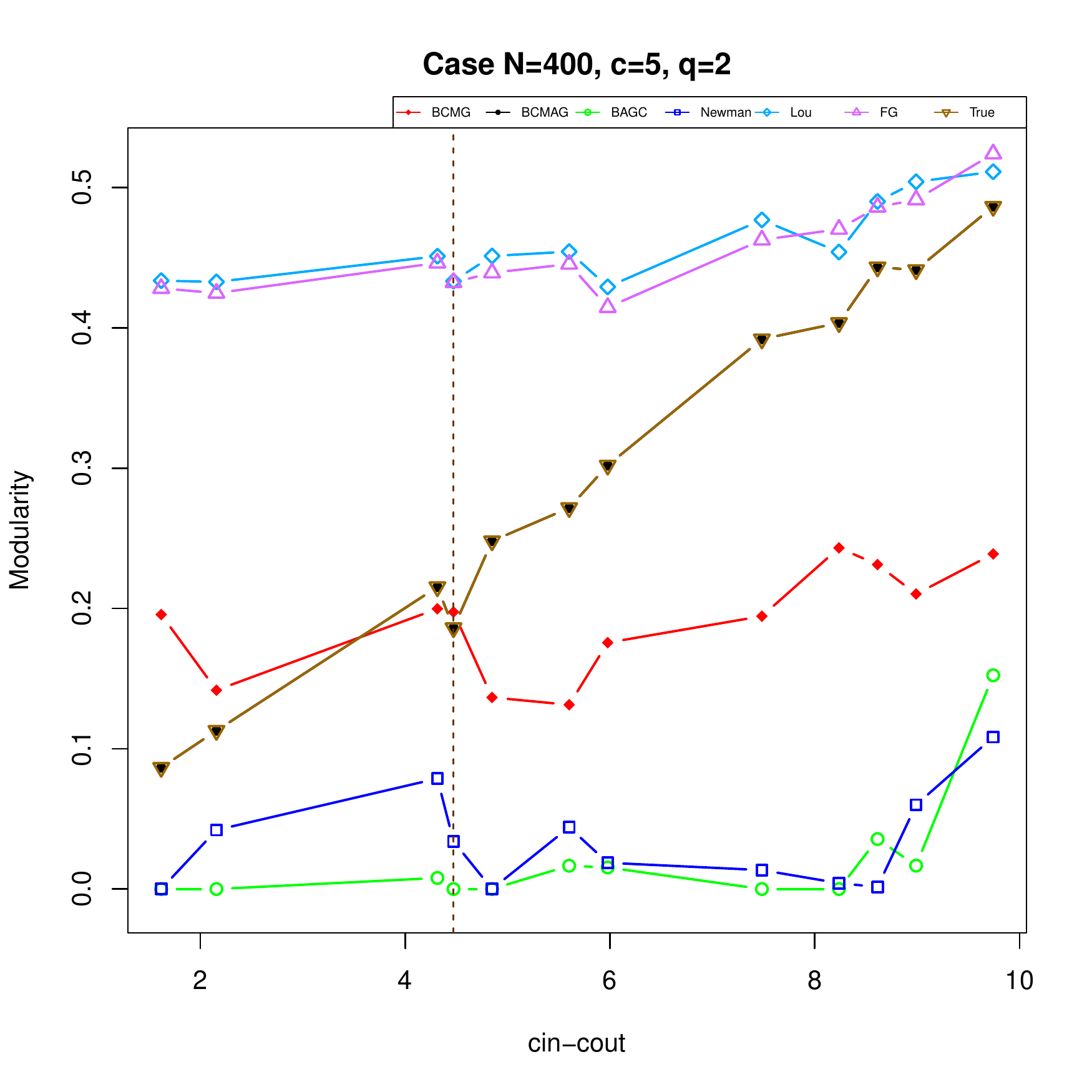}
            \caption[]%
            {{\small}}
            \label{fig:Qq=2c=5}
        \end{subfigure}
        \hfill
        \begin{subfigure}[b]{0.485\textwidth}
            \centering
            \includegraphics[page=1, width=\textwidth]{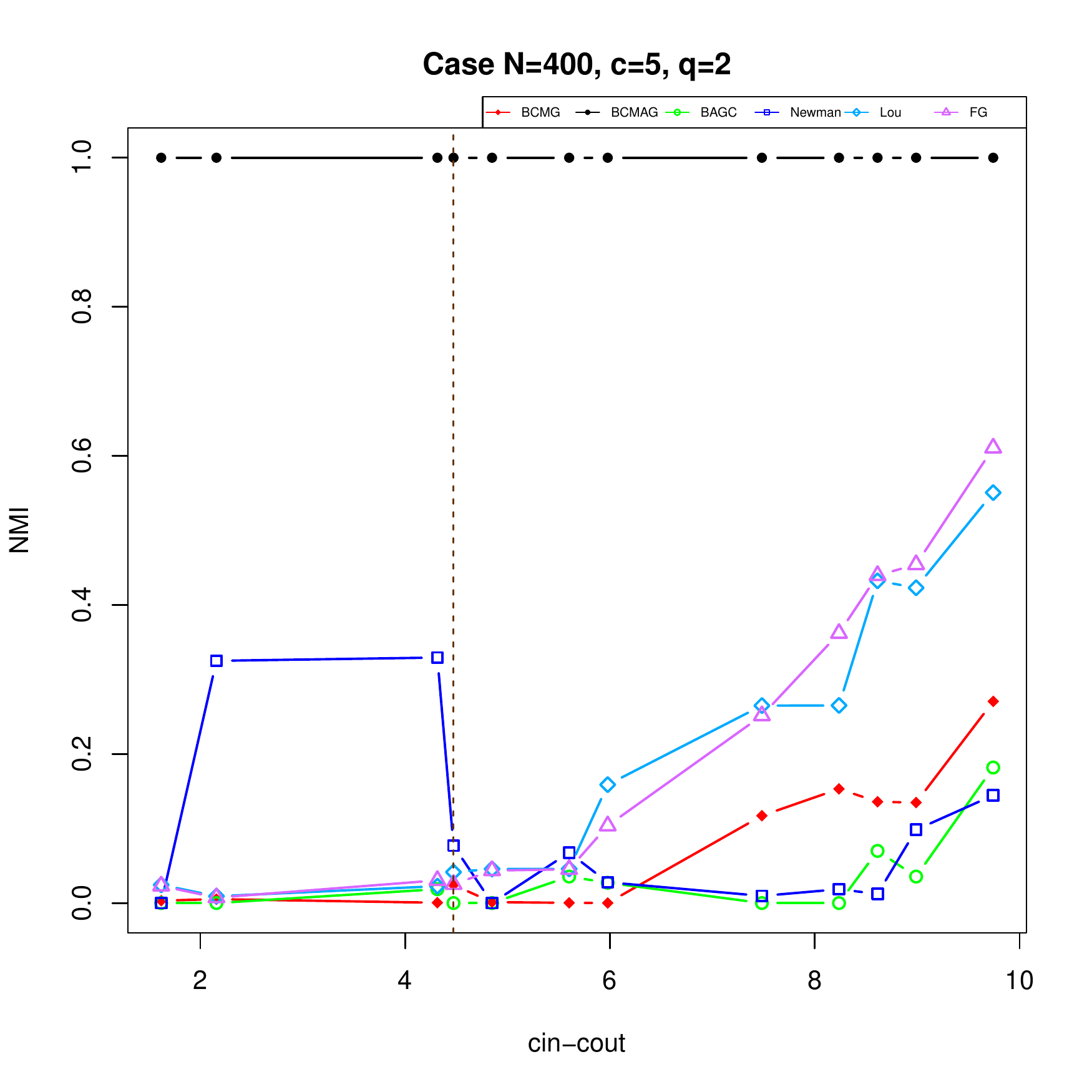}
            \caption[]%
            {{\small }}
            \label{fig:nmiq=2c=5}
        \end{subfigure}
         \vspace{.1\baselineskip}
        \begin{subfigure}[b]{0.485\textwidth}
            \centering
            \includegraphics[page=1, width=\textwidth]{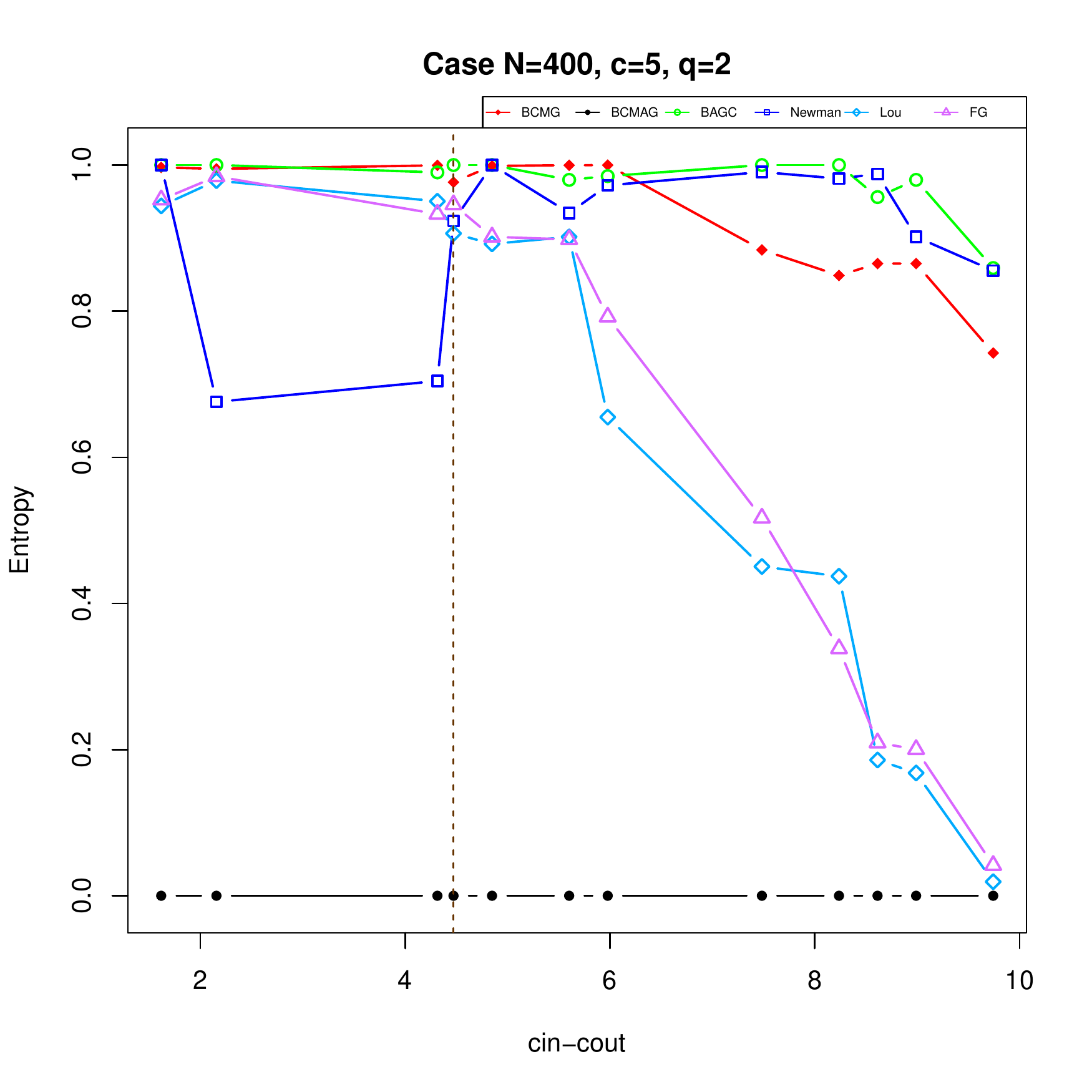}
            \caption[]%
            {{\small }}
            \label{fig:Eq=2c=5}
        \end{subfigure}
        \hfill
        \begin{subfigure}[b]{0.485\textwidth}
            \centering
            \includegraphics[page=1, width=\textwidth]{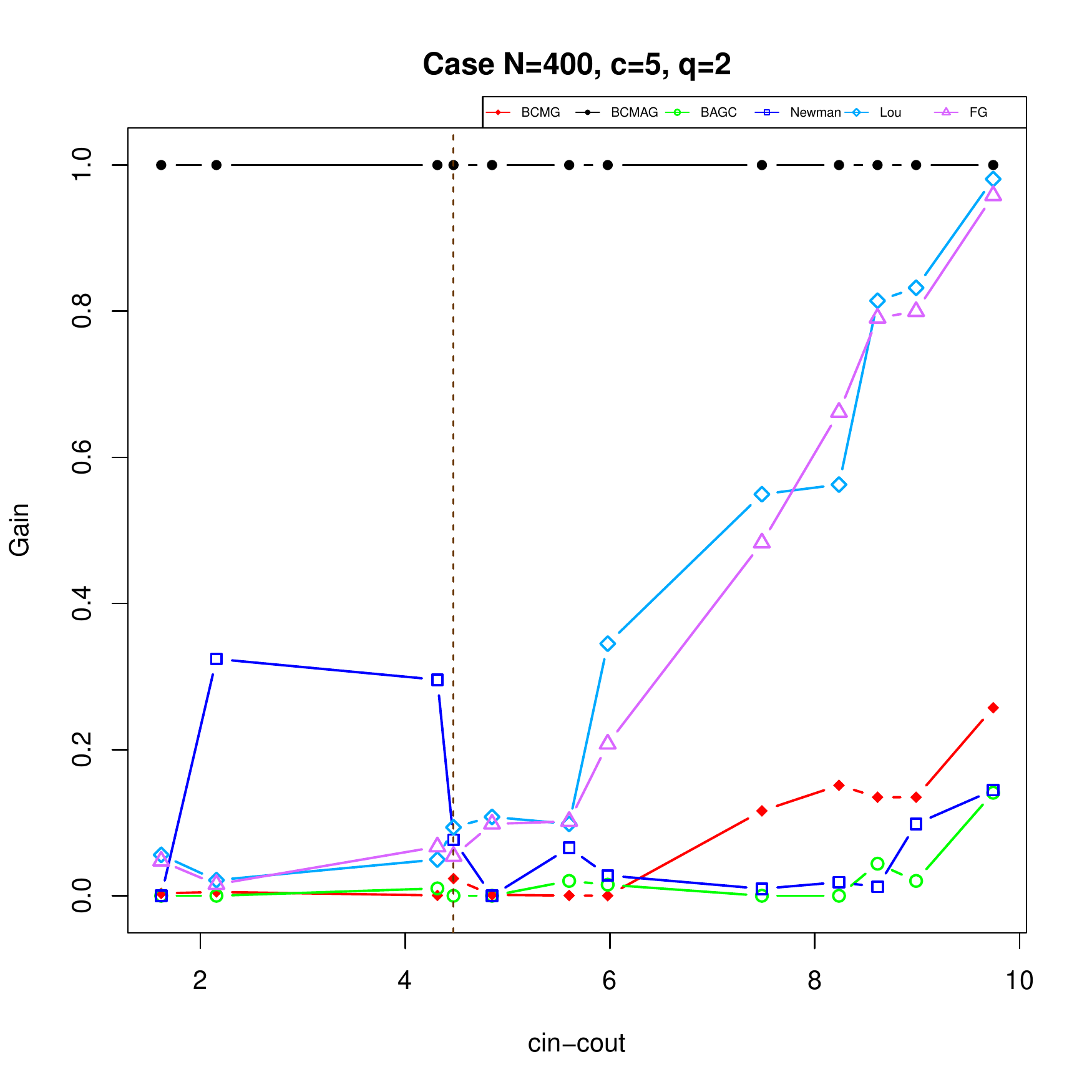}
            \caption[]%
            {{\small }}
            \label{fig:Gq=2c=5}
        \end{subfigure}
        \caption[]
       {\small SBM dataset. Case $n=400$, $q=2$, $c=5$. In each subplot we indicated the theoretical detectability threshold with a vertical dashed line, and modularity (a), $NMI$ (b), entropy (c) and gain (d) corresponding to the different clustering approaches.}
        \label{fig:Compq=2c=5}
    \end{figure*}

\begin{figure*}[h]
        \centering
        \begin{subfigure}[b]{0.485\textwidth}
            \centering
            \includegraphics[page=4, width=\textwidth]{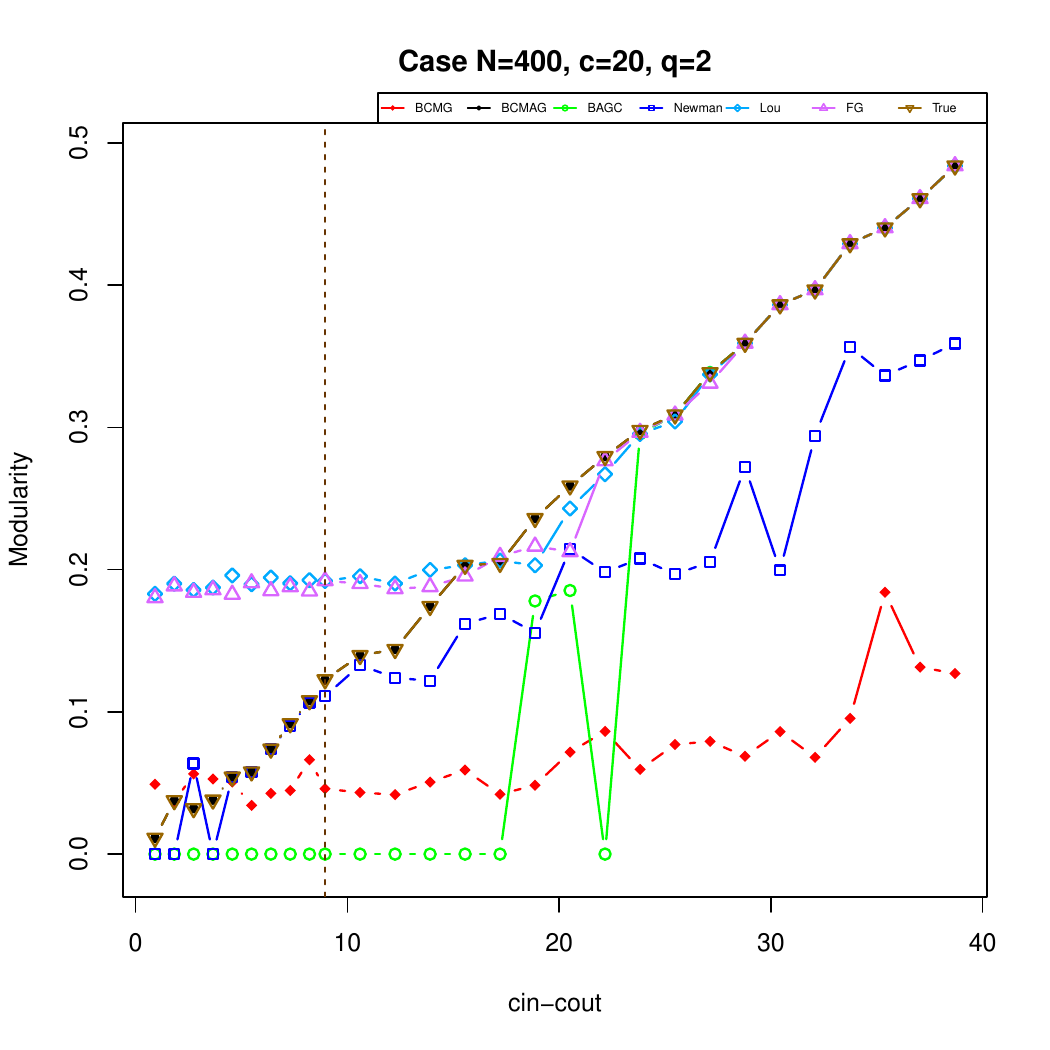}
            \caption[]%
            {{\small}}
            \label{fig:Qq=2c=20}
        \end{subfigure}
        \hfill
        \begin{subfigure}[b]{0.485\textwidth}
            \centering
            \includegraphics[page=4, width=\textwidth]{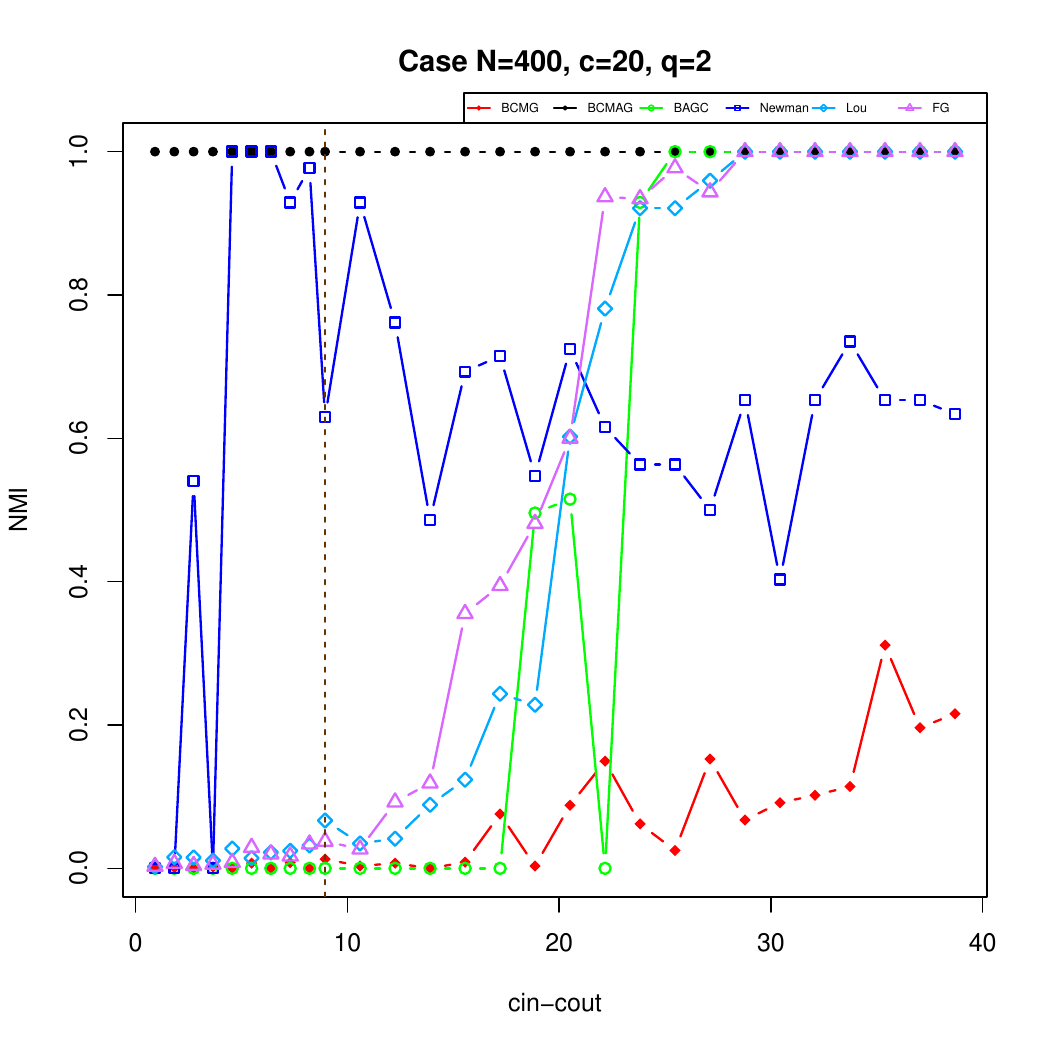}
            \caption[]%
            {{\small }}
            \label{fig:nmiq=2c=20}
        \end{subfigure}
         \vspace{.1\baselineskip}
        \begin{subfigure}[b]{0.485\textwidth}
            \centering
            \includegraphics[page=4, width=\textwidth]{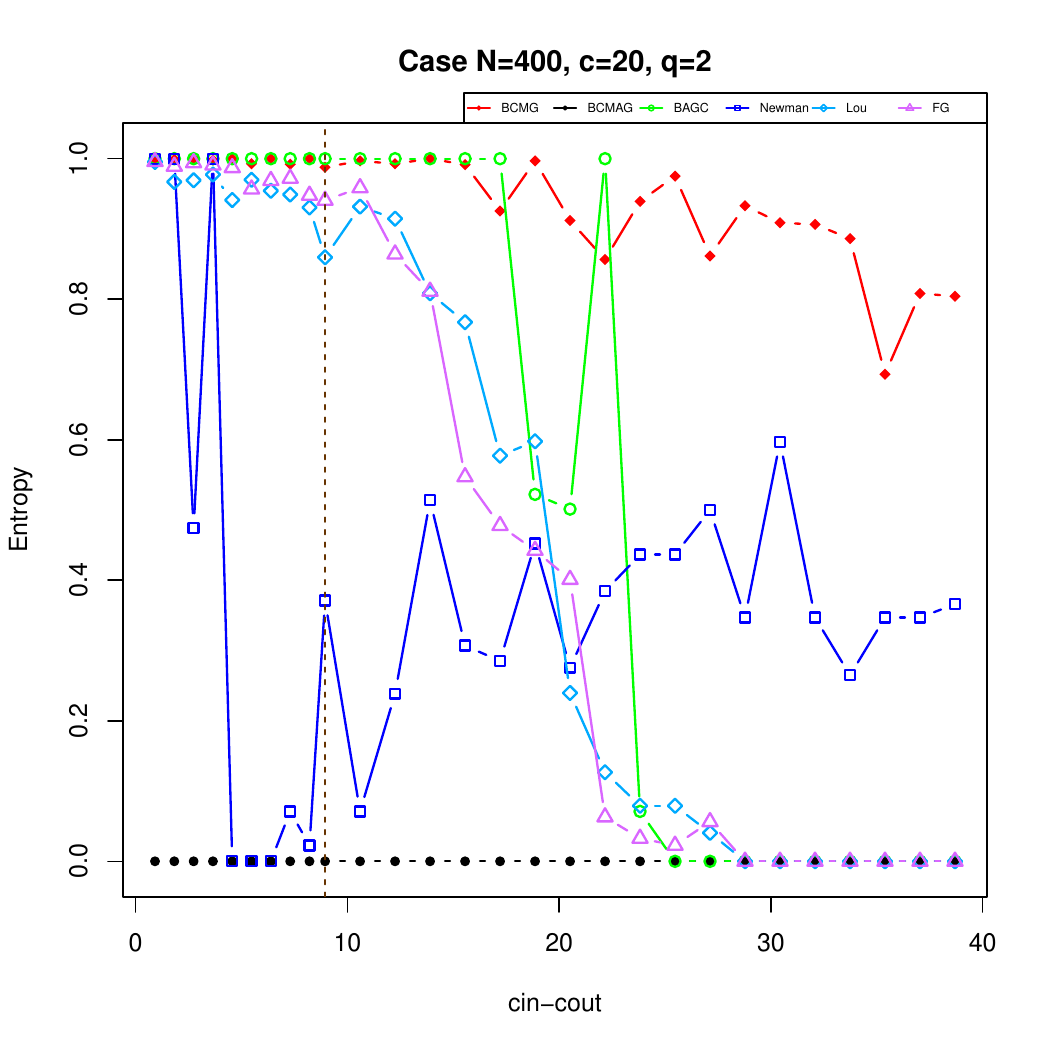}
            \caption[]%
            {{\small }}
            \label{fig:Eq=2c=20}
        \end{subfigure}
        \hfill
        \begin{subfigure}[b]{0.485\textwidth}
            \centering
            \includegraphics[page=4, width=\textwidth]{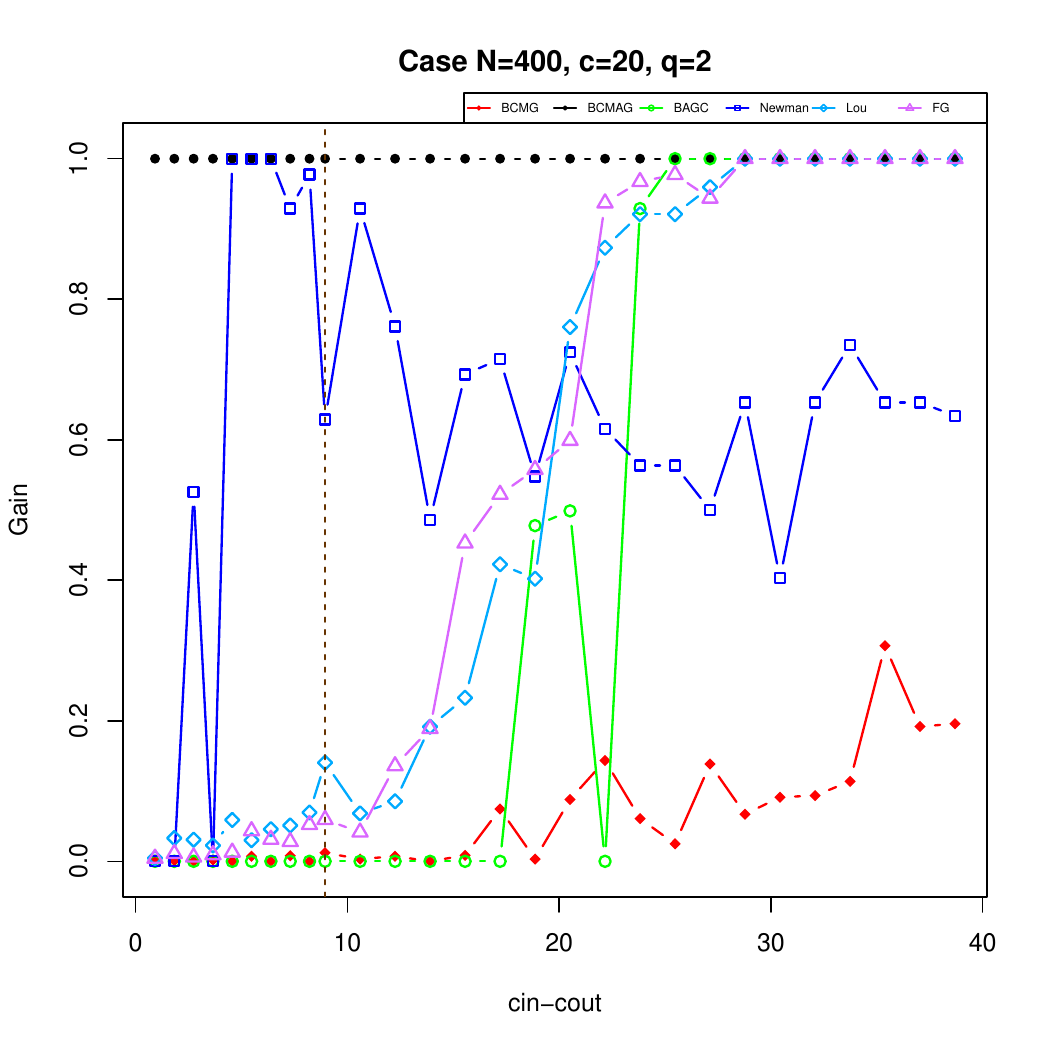}
            \caption[]%
            {{\small }}
            \label{fig:Gq=2c=20}
        \end{subfigure}
        \caption[]
       {\small SBM dataset. Case $n=400$, $q=2$, $c=20$. In each subplot we indicated the theoretical detectability threshold with a vertical dashed line, and modularity (a), $NMI$ (b), entropy (c) and gain (d) corresponding to the different clustering approaches.}
        \label{fig:Compq=2c=20}
    \end{figure*}

\begin{figure*}[h]
        \centering
        \begin{subfigure}[b]{0.485\textwidth}
            \centering
            \includegraphics[page=9, width=\textwidth]{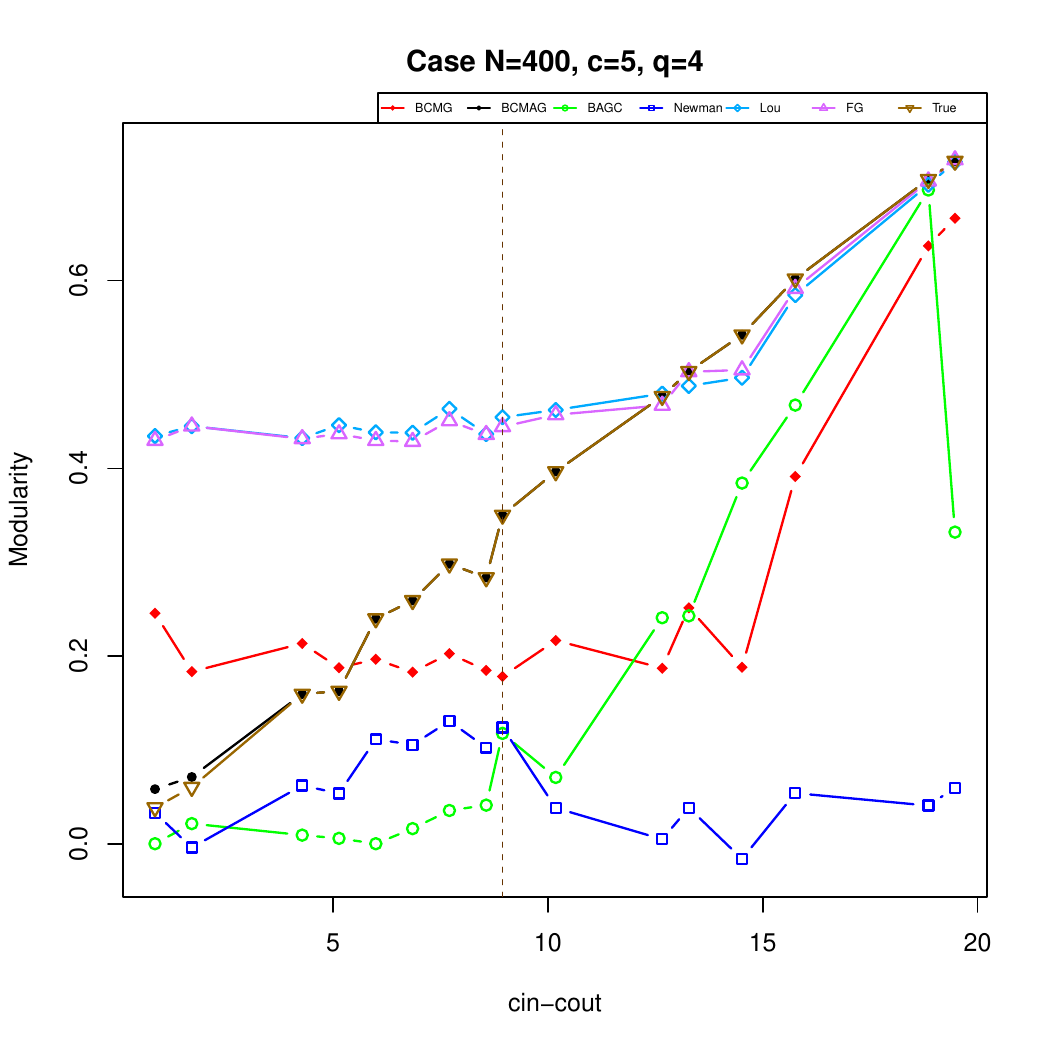}
            \caption[]%
            {{\small}}
            \label{fig:Qq=4c=5}
        \end{subfigure}
        \hfill
        \begin{subfigure}[b]{0.485\textwidth}
            \centering
            \includegraphics[page=9, width=\textwidth]{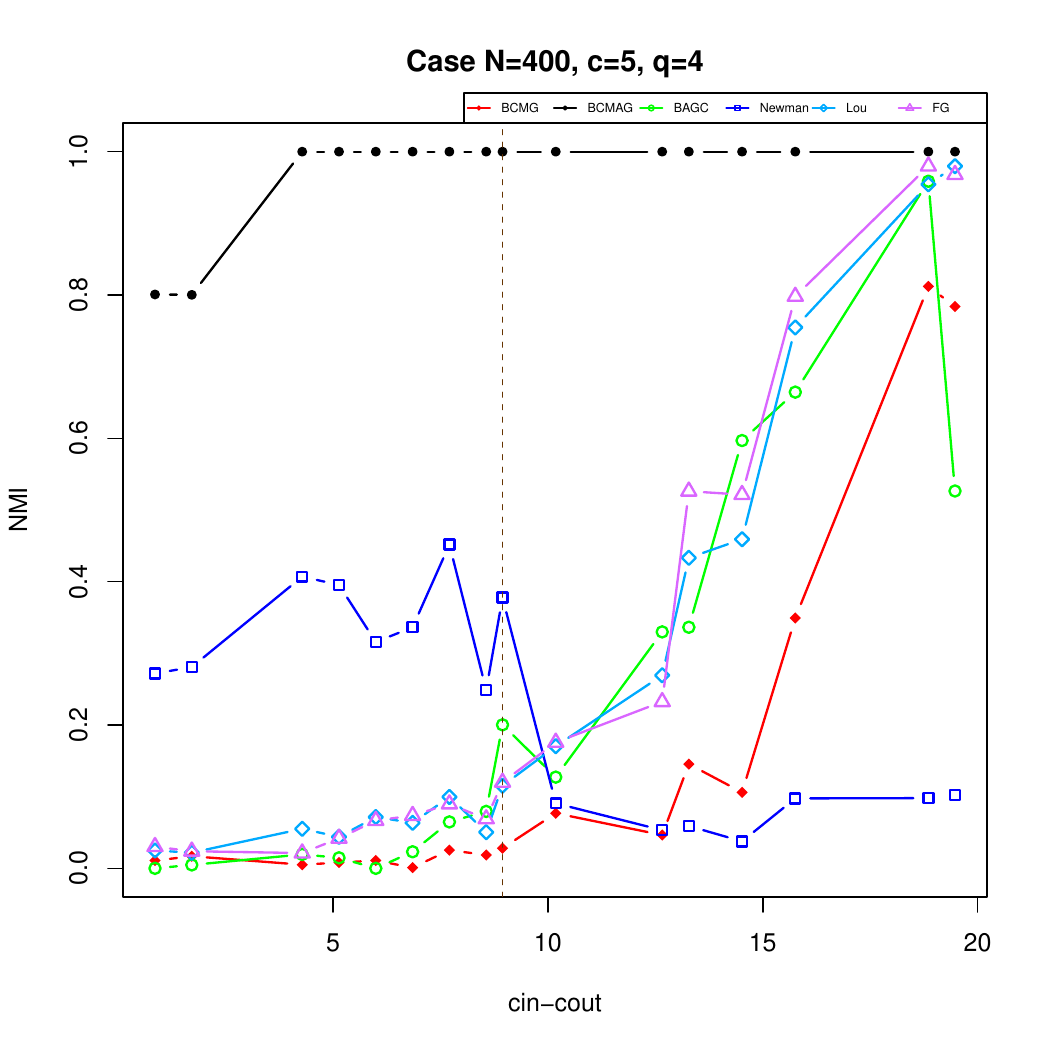}
            \caption[]%
            {{\small }}
            \label{fig:nmiq=4c=5}
        \end{subfigure}
         \vspace{.1\baselineskip}
        \begin{subfigure}[b]{0.485\textwidth}
            \centering
            \includegraphics[page=9, width=\textwidth]{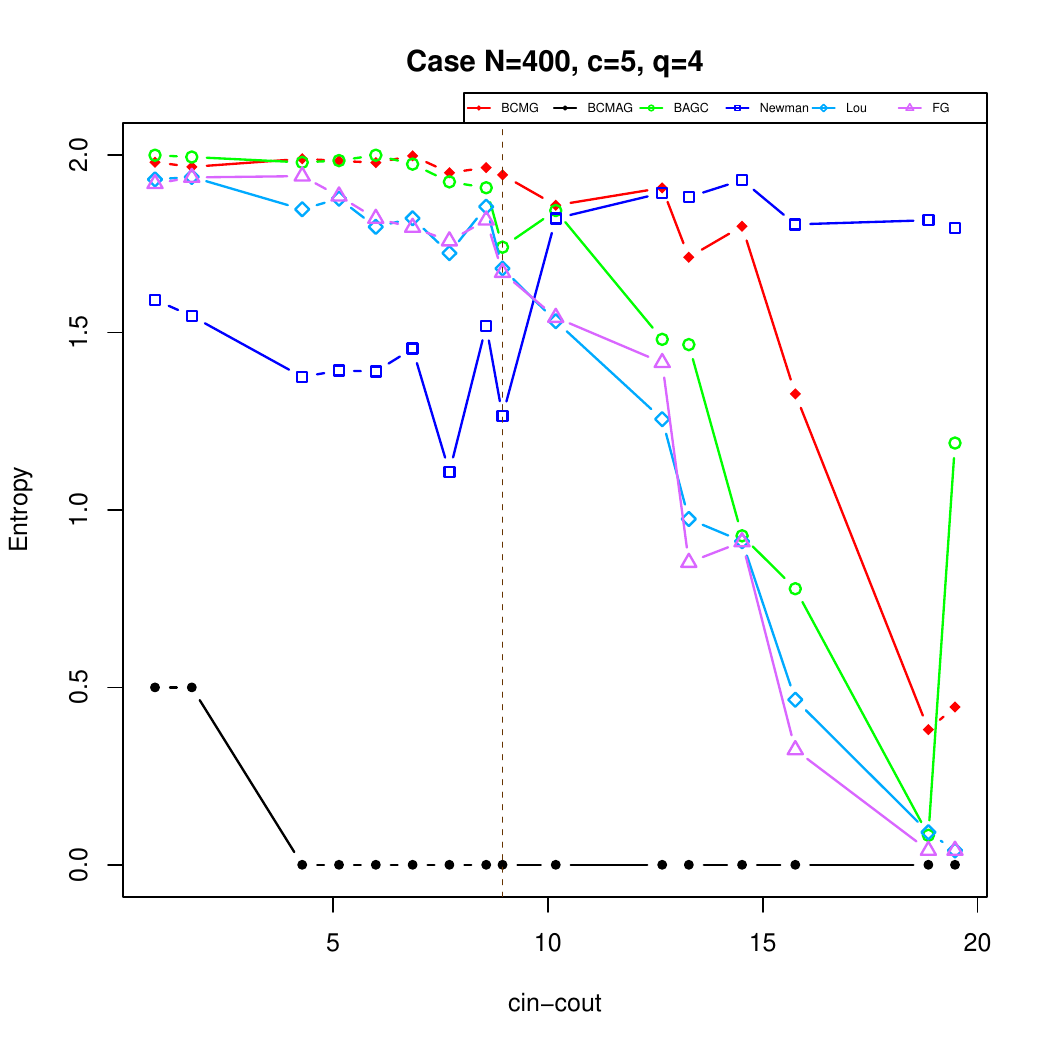}
            \caption[]%
            {{\small }}
            \label{fig:Eq=4c=5}
        \end{subfigure}
        \hfill
        \begin{subfigure}[b]{0.485\textwidth}
            \centering
            \includegraphics[page=9, width=\textwidth]{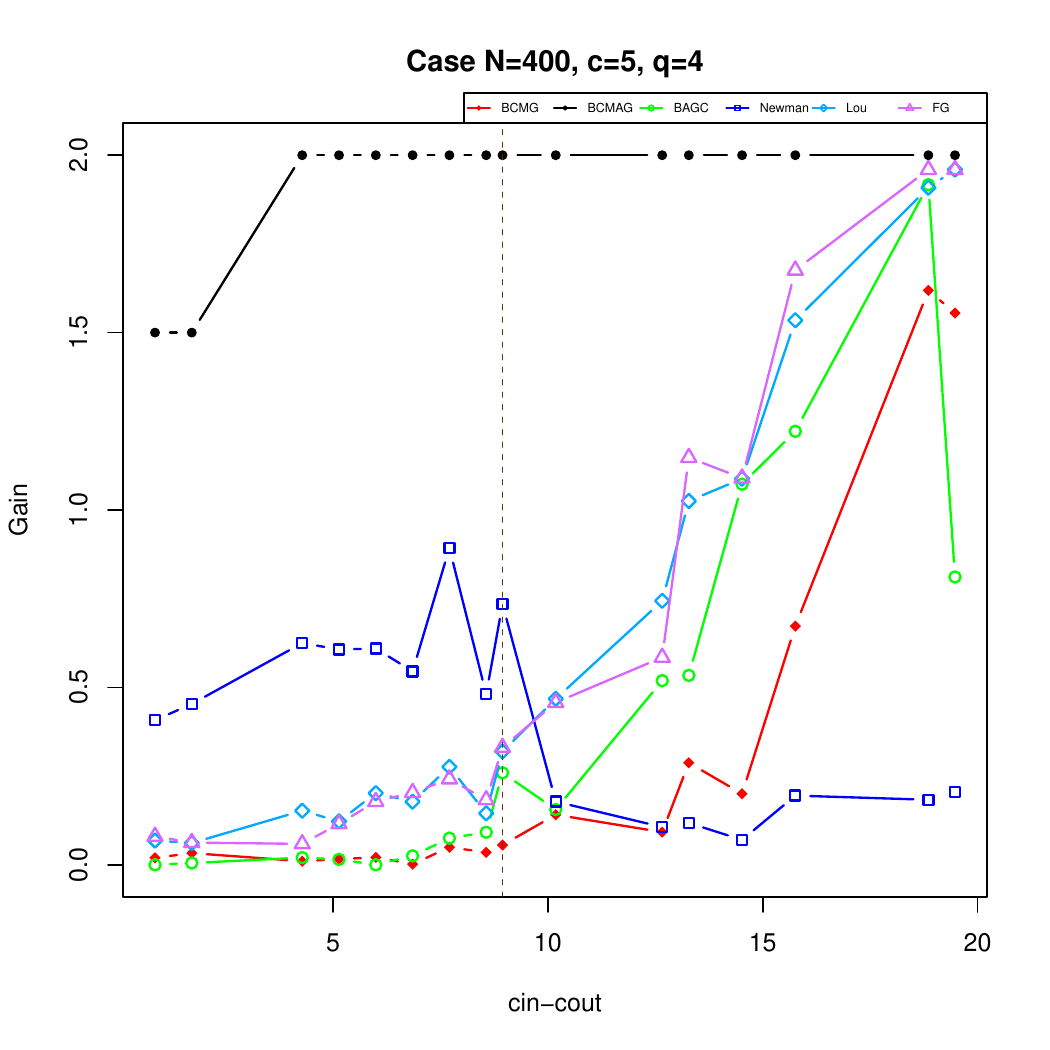}
            \caption[]%
            {{\small }}
            \label{fig:Gq=4c=5}
        \end{subfigure}
        \caption[]
       {\small SBM dataset. Case $n=400$, $q=4$, $c=5$. In each subplot we indicated the theoretical detectability threshold with a vertical dashed line, and modularity (a), $NMI$ (b), entropy (c) and gain (d) corresponding to the different clustering approaches.}
        \label{fig:Compq=4c=5}
    \end{figure*}

\begin{figure*}[h]
        \centering
        \begin{subfigure}[b]{0.485\textwidth}
            \centering
            \includegraphics[page=12, width=\textwidth]{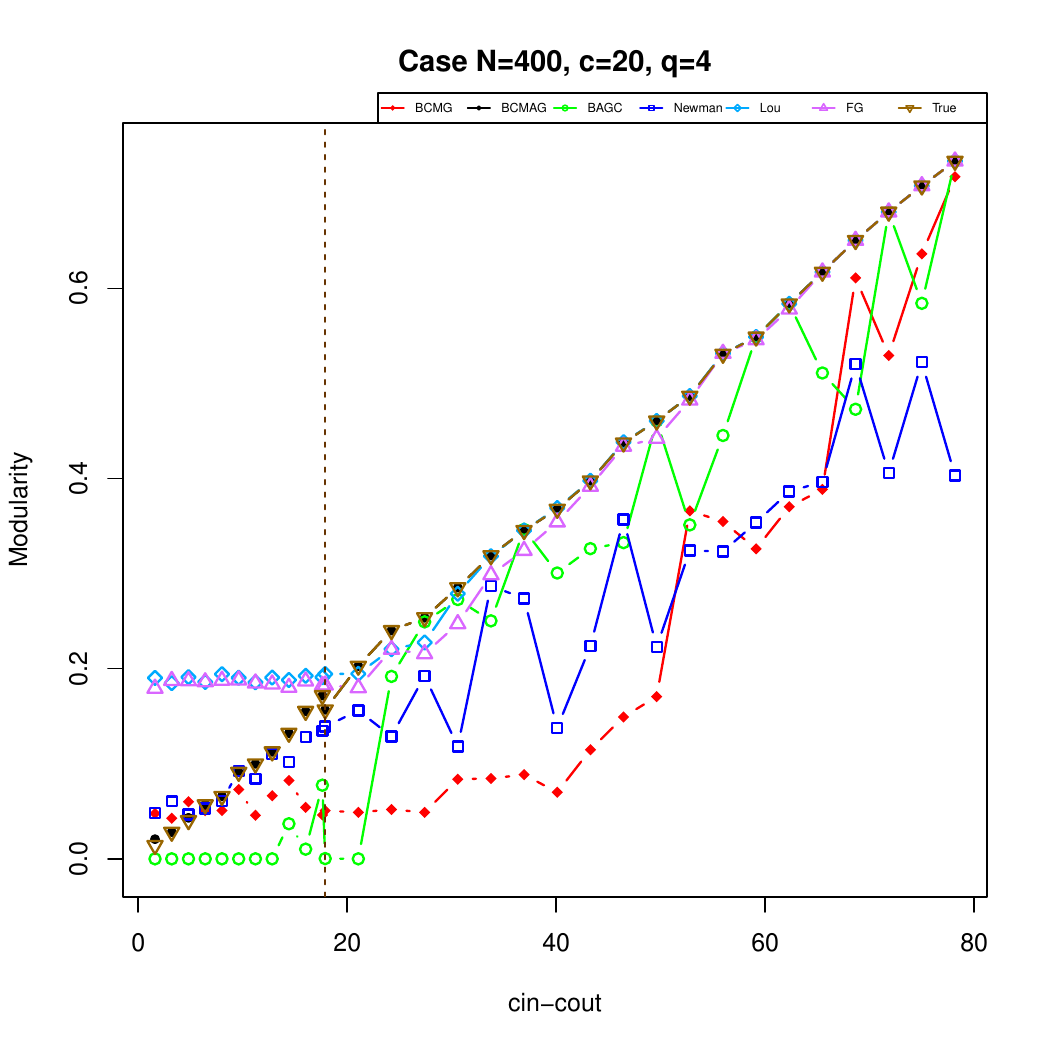}
            \caption[]%
            {{\small}}
            \label{fig:Qq=4c=20}
        \end{subfigure}
        \hfill
        \begin{subfigure}[b]{0.485\textwidth}
            \centering
            \includegraphics[page=12, width=\textwidth]{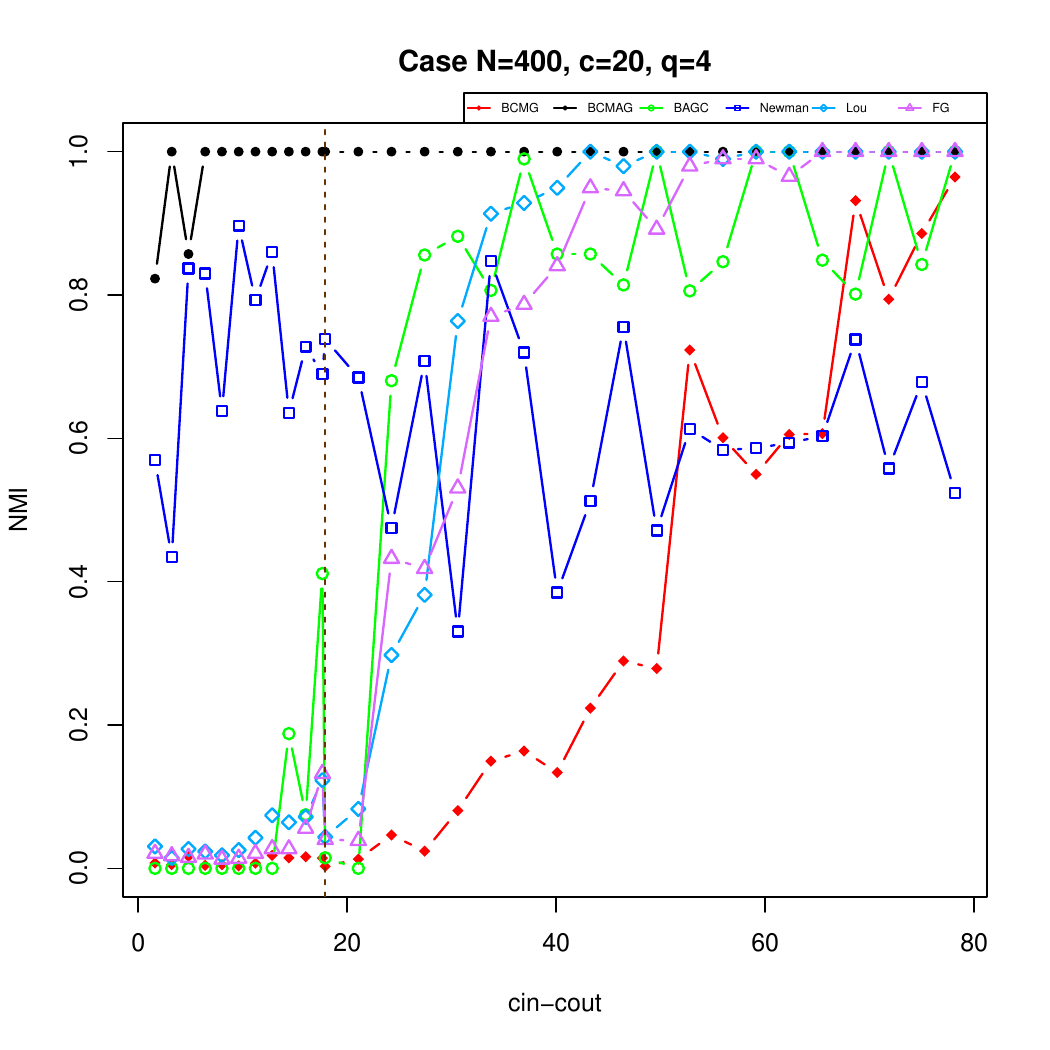}
            \caption[]%
            {{\small }}
            \label{fig:nmiq=4c=20}
        \end{subfigure}
         \vspace{.1\baselineskip}
        \begin{subfigure}[b]{0.485\textwidth}
            \centering
            \includegraphics[page=12, width=\textwidth]{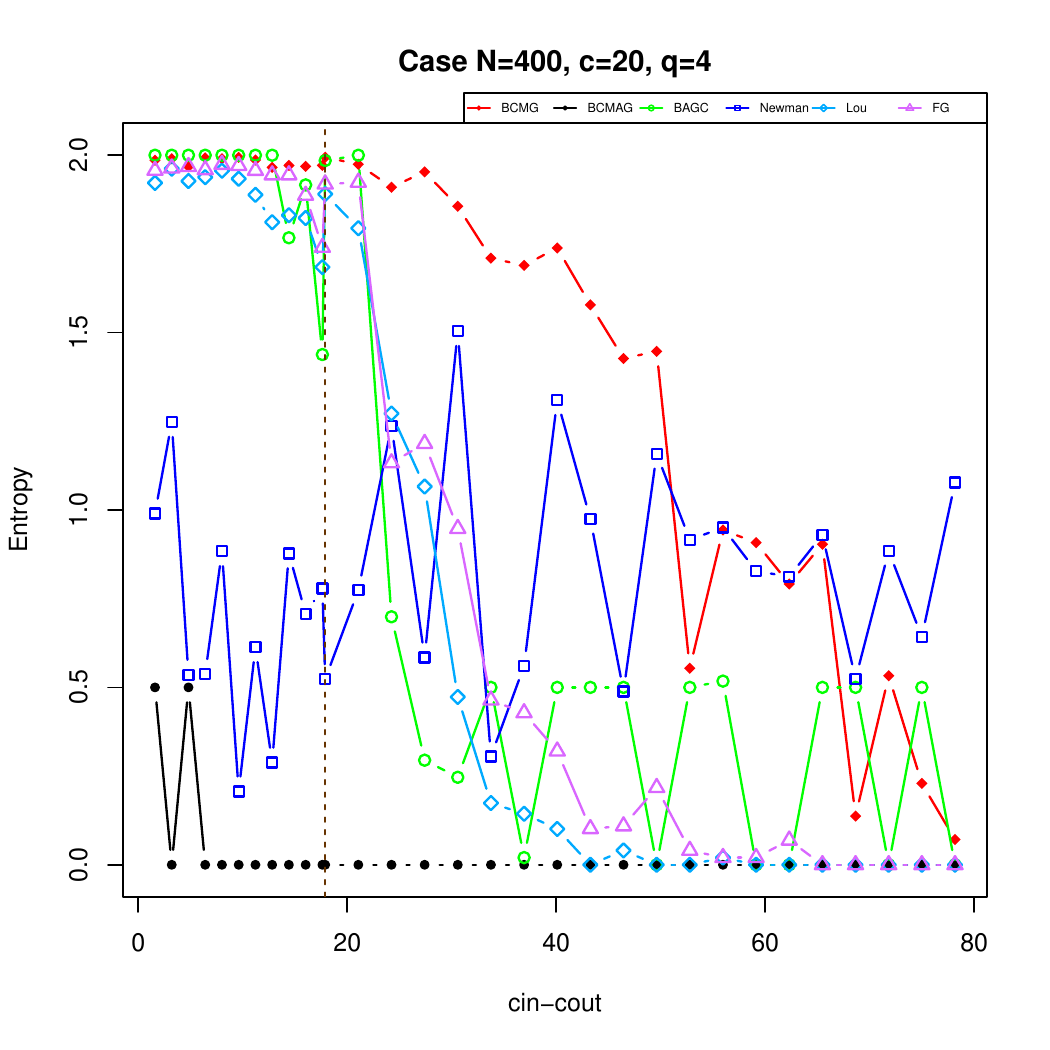}
            \caption[]%
            {{\small }}
            \label{fig:Eq=4c=20}
        \end{subfigure}
        \hfill
        \begin{subfigure}[b]{0.485\textwidth}
            \centering
            \includegraphics[page=12, width=\textwidth]{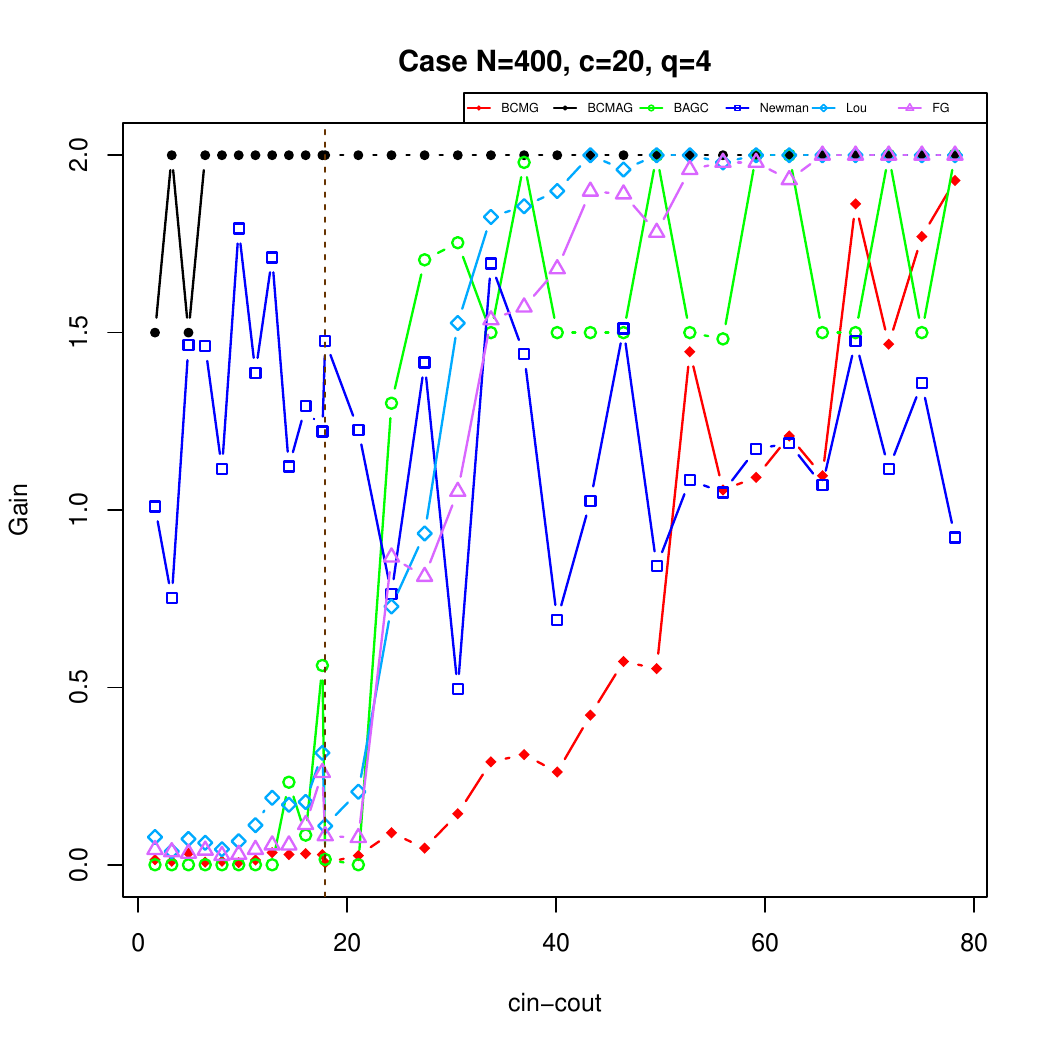}
            \caption[]%
            {{\small }}
            \label{fig:Gq=4c=20}
        \end{subfigure}
        \caption[]
       {\small SBM dataset. Case $n=400$, $q=4$, $c=20$. In each subplot we indicated the theoretical detectability threshold with a vertical dashed line, and modularity (a), $NMI$ (b), entropy (c) and gain (d) corresponding to the different clustering approaches.}
        \label{fig:Compq=4c=20}
    \end{figure*}

In Figures~\ref{fig:Compq=2c=5}-\ref{fig:Compq=4c=20}, we show a summary of the results obtained for $n=400$ vertices, $q=2,4$, and $c=5,20$. In our simulated data the ground truth labels are known and we plot the corresponding modularity in subfigure (a) of every plot. As you can see, our new method $BCMAG$ perfectly matches the truth also below the detectability threshold, especially for $q=2$. Furthermore, $BCMAG$ largely outperforms  $BCMG$, at higher vertex degree and graph density (i.e., for $c=20$). On the other hand, it is expected that a method that does not use the attributes would fail below the detectability threshold. The two methods $FG$ and $Lou$ do not use labels and maximize the modularity and they both show an overall overestimation of the modularity in the case of sparse graphs ($c=5$). If we look at the $NMI$, entropy and gain values, we can see that in all cases for $q=2$, our new method $BCMAG$ performs perfectly well below the detectability threshold, while for $q=4$, some small discrepancy is observed when $c_{in}-c_{out}$ is very small (i.e., when modularity is very small). This analysis shows that $BCMAG$ has very good ability to detect clusters which are correlated to planted SBM synthetic graphs also when the vertex groups are not clearly formed and the graph structure is ambiguous. It is worth noticing that $NMI$ and entropy values for both $Newman$ and $BAGC$, are not satisfactory especially for sparser graphs ($c=5$) although both methods make use of the attributes information.

\subsection{LFR-EA graphs}
\label{sub:LFR}

In this section we discuss results obtained on the synthetic graphs introduced in~\cite{EA2013} as reference benchmark to assess performance of clustering algorithms using vertex attributes information.
This benchmark is based on the Lancichinetti–Fortunato–Radicchi (LFR) model~\cite{LFR2008}, which uses power-law distributions for both vertex degree and community size. This model generates more realistic networks with respect to the above SBM, providing heterogeneity in the distributions of both node degrees and of community sizes.
A {\em structure mixing} parameter ranging in $[0,1]$ controls the fraction of edges that are between communities, i.e., the block structure in the network. If the structure mixing is equal to $0$, the network has a well partitioned structure, i.e., it has a large modularity. On the other hand, when the structure mixing parameter approaches $1$, the structure becomes less modular. In~\cite{EA2013}, the authors introduce a model to generate attributes for the LFR graph vertices. The model parameters are the number $m$ of attributes $l_j, \ j=1, \ldots, m$, the domain size for the attribute values $n_j$ and the {\em attributes noise} level.  All the vertices in a cluster are assumed to share the same attribute domain values. The attributes of some randomly chosen vertices in each cluster are assigned attribute values affected by the specified noise level, so that this benchmark model can be used to analyze the robustness of the algorithms with respect to perturbations in the attribute values.
This benchmark model, named LFR-EA, is often used in the literature~\cite{PS2020,BMFM2022} to compare results with those discussed in~\cite[section 5.2]{EA2013} also because several methods implementations are not made available by the authors.

\begin{figure*}[htb]
        \centering
        \begin{subfigure}[b]{0.4\textwidth}
            \centering
            \includegraphics[width=\textwidth]{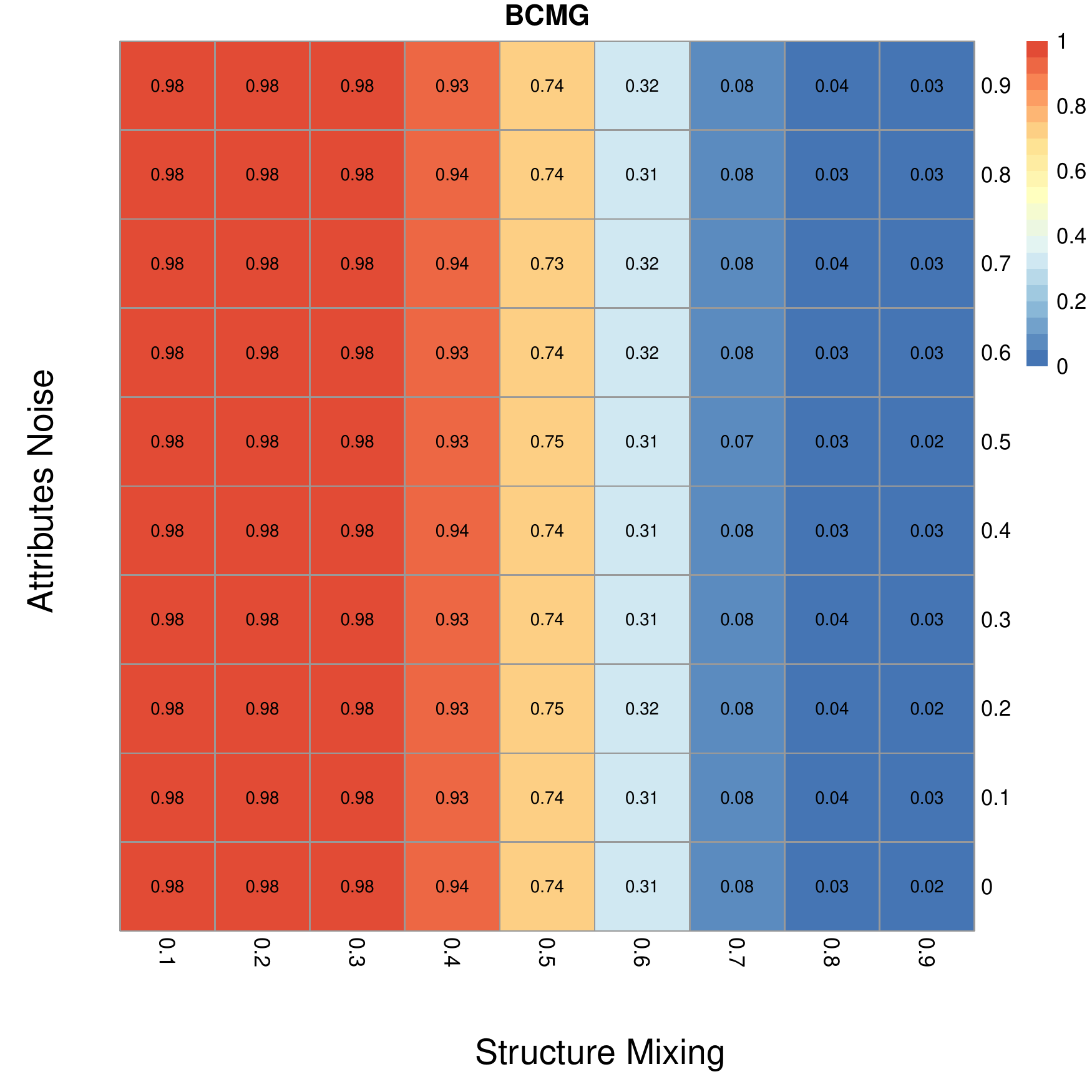}
            \label{fig:heatbcmg}
        \end{subfigure}
        \begin{subfigure}[b]{0.4\textwidth}
            \centering
            \includegraphics[width=\textwidth]{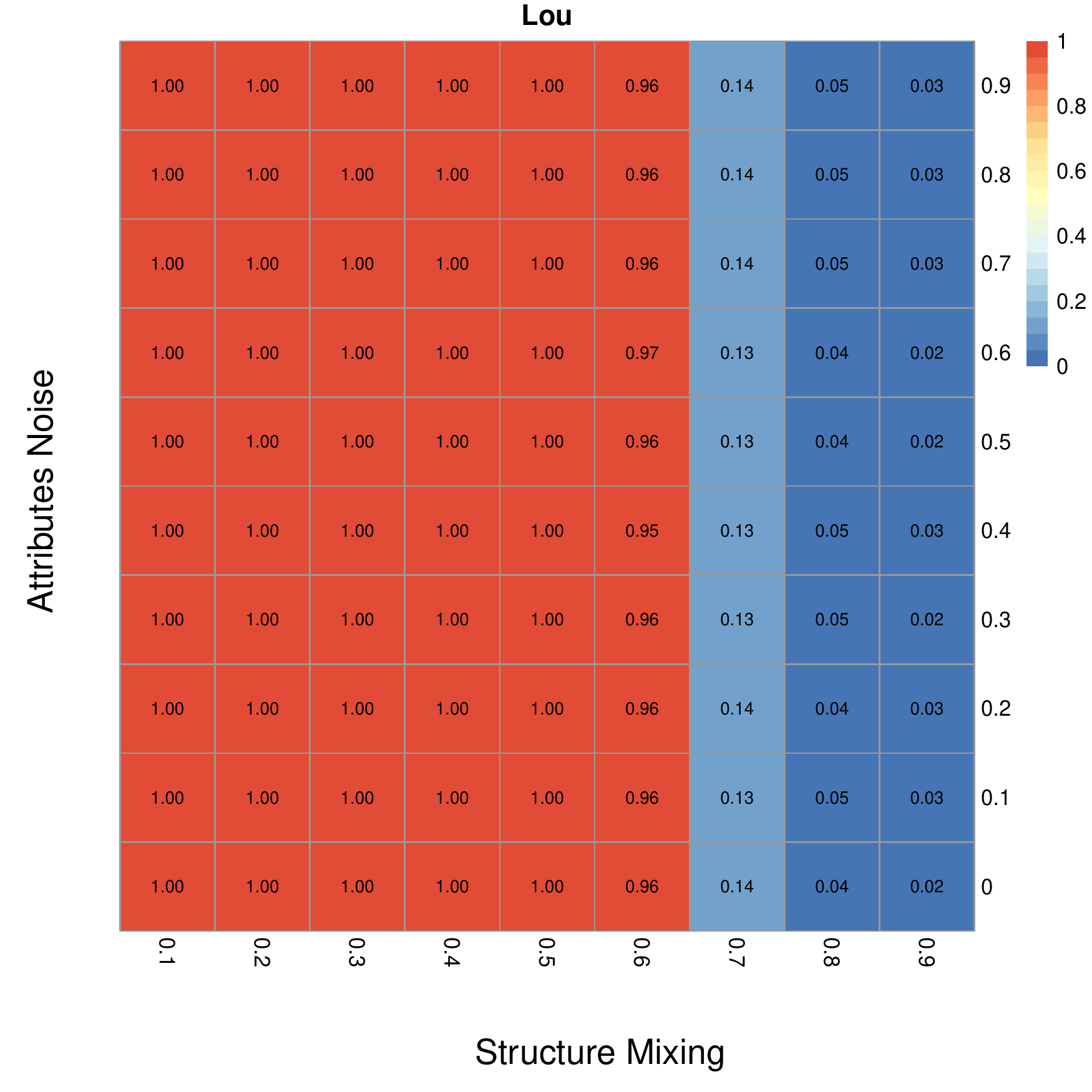}
            \label{fig:heatlou}
        \end{subfigure}
        \begin{subfigure}[b]{0.4\textwidth}
            \centering
            \includegraphics[width=\textwidth]{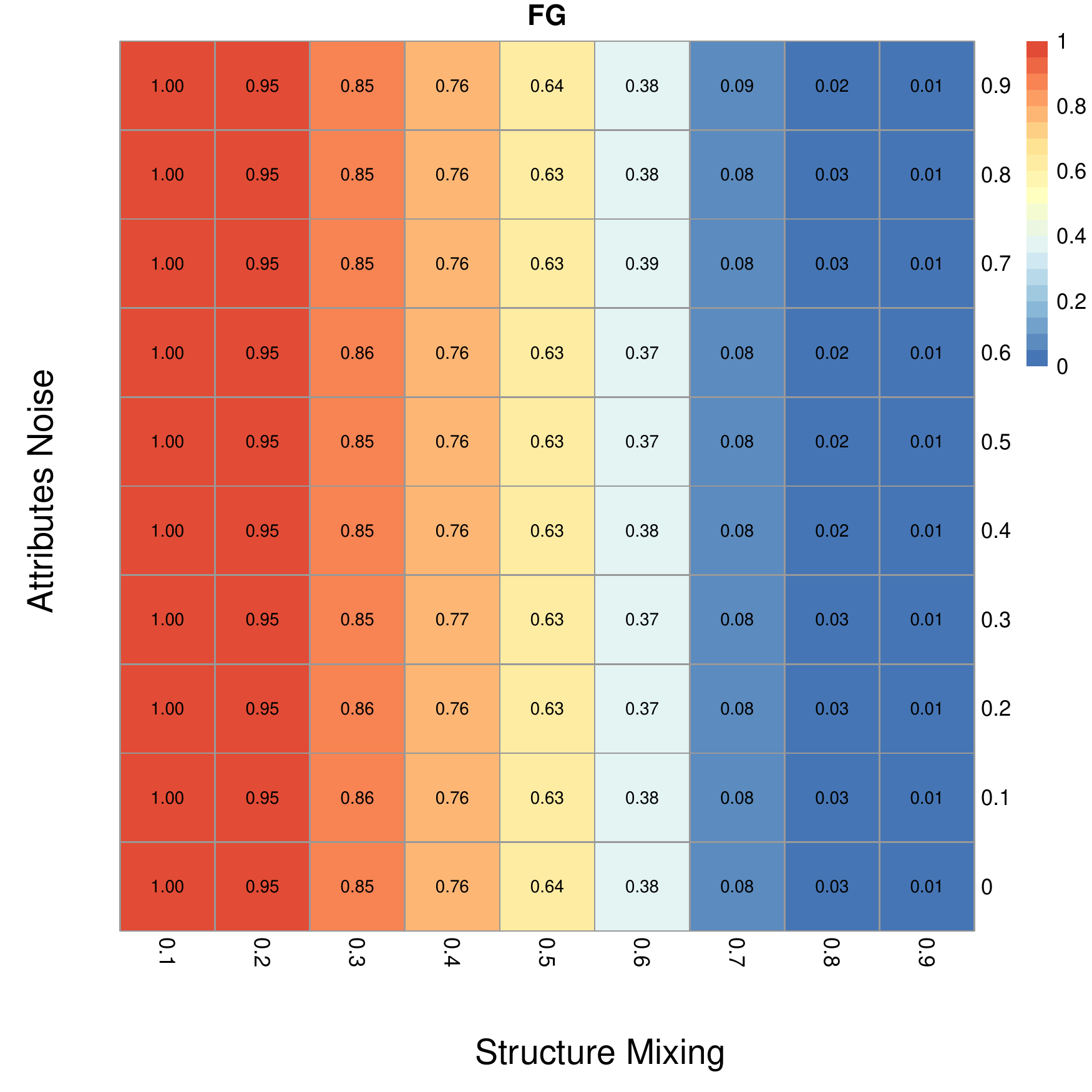}
            \label{fig:heatfg}
        \end{subfigure}
        \vspace{.1\baselineskip}
       \begin{subfigure}[b]{0.4\textwidth}
            \centering
            \includegraphics[width=\textwidth]{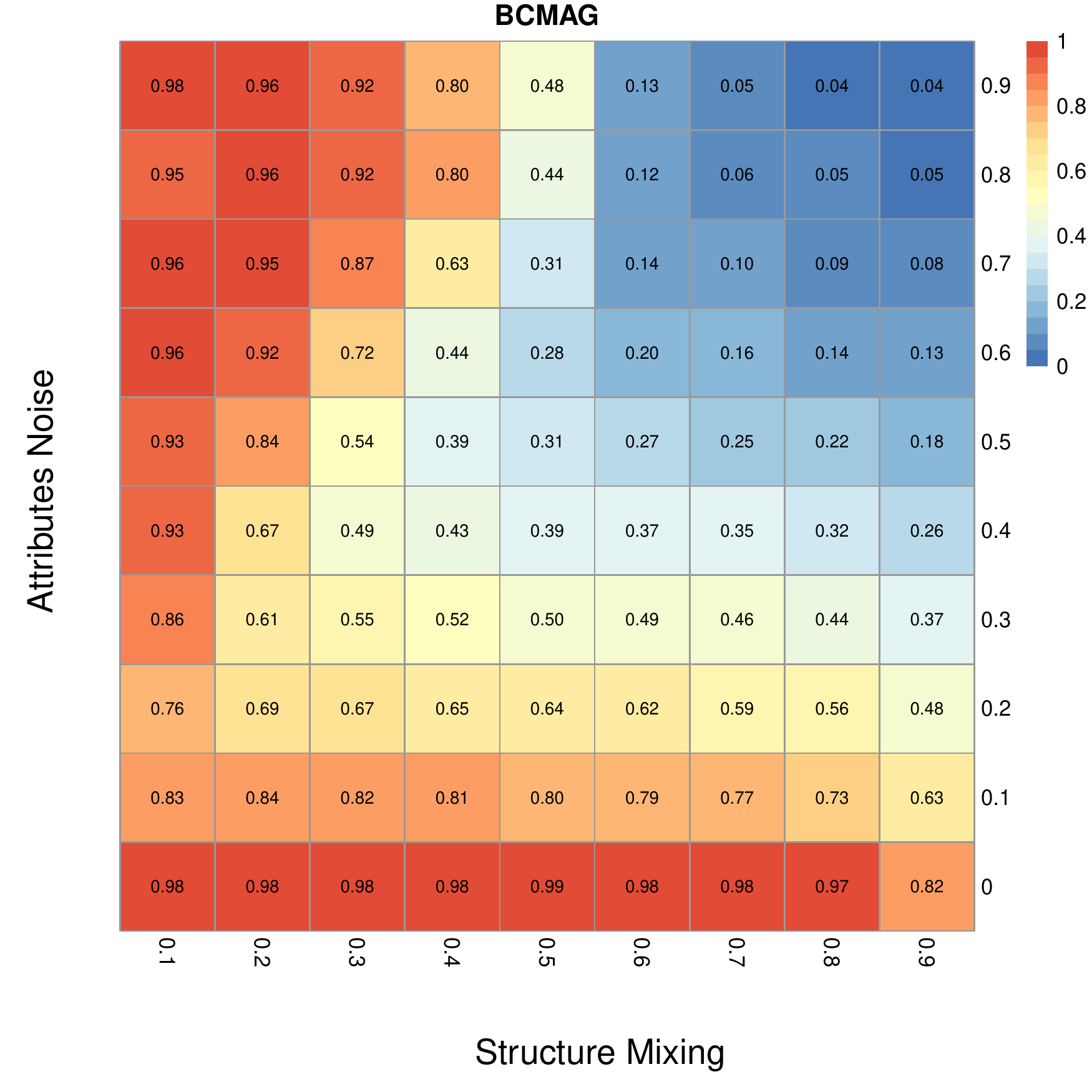}
            \label{fig:heatbcmag}
        \end{subfigure}
        \begin{subfigure}[b]{0.4\textwidth}
                \centering
            \includegraphics[width=\textwidth]{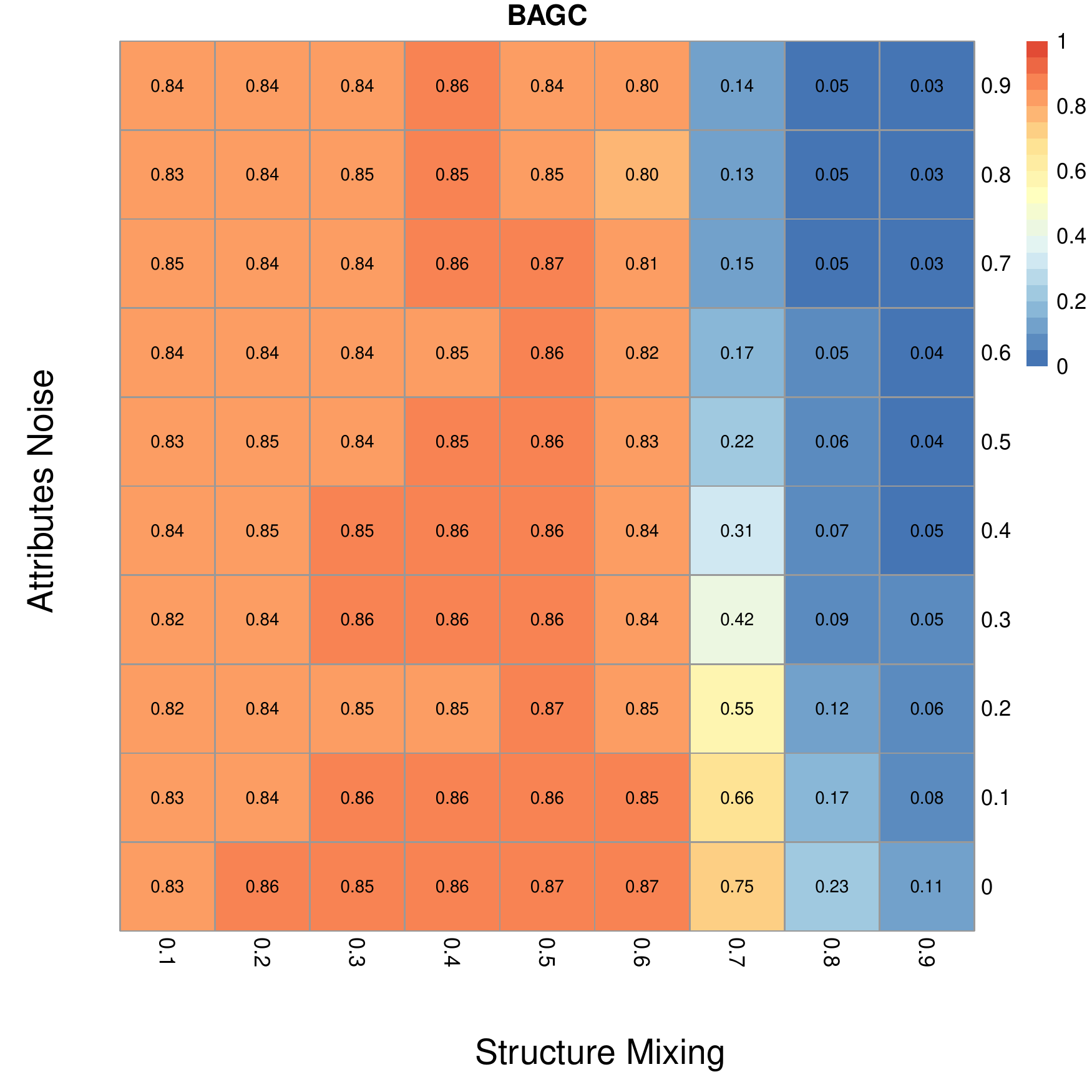}
            \label{fig:heatbagc}
        \end{subfigure}
        \vspace{-0.3cm}
         \caption[]
       {\small LFR-EA dataset. Each subplot illustrates the NMI heatmaps for each method; the color range is based on mean NMI for $100$ graph samples. The x-axis represents the structure mixing parameter and the y-axis represents the attributes noise. $BCMG$ (a), $Lou$ (b), $FG$ (c), $BCMAG$ (d) and $BAGC$ (e).}
\label{fig:lfr-ea}
\end{figure*}

We generated LFR-EA with the set of parameters detailed in~\cite[Table 1]{EA2013}. The generated graphs have $1000$ vertices and $2$ classes of attributes. Following the same analysis approach in~\cite{EA2013}, we considered 9 structure mixing parameters $[0.1,0.2,\ldots, 0.9]$, and 10 attribute noise levels $[0,0.1,\ldots, 0.9]$. For each couple of mixing parameter and attribute noise level, we generated $100$ samples graphs, for a total of $9000$ test cases.
On each attributed graph, we run a set of clustering methodologies and  we summarised the results in terms of $NMI$ heatmaps in Figure~\ref{fig:lfr-ea}. Note that each color represents the mean $NMI$ value corresponding to $100$ graph samples generated in correspondence of the specific  set of mixing parameter and attribute noise levels. In each heatmap the x-axis represents the structure mixing parameter and the y-axis represents the attributes noise.

As expected, the clustering algorithms which only use information on the structure, such as $BCMG$, $Lou$ and $FG$, show the same values of $NMI$ in vertical columns, indeed attributes have no impact on the performance of those algorithms. As in the SBM-based graph samples, $BCMG$ confirms his ability to well identify partitions in highly modular graphs also in the case of LFR benchmark, and shows better accuracy than $FG$ in recovering the clusters for decreasing graph modularity. Furthermore, as well known, $Lou$ has the best performance with a perfect $NMI$ ($\sim 1$) as long as a structure mixing smaller or equal to $0.5$ is used, and it shows very high values of mean $NMI$ also when the structure mixing reaches the value of $0.6$. However, when structure mixing parameter increases, all methods are unable to recover correctly the clusters.

If we look at the average NMI of the methods employing information on attributes, we observe that increasing attributes noise has more impact on $BCMAG$ than onto $BACG$, especially when increasing structure mixing parameter. On the other hand, the top-left and the bottom-right corners of the $BCMAG$ subplot square show it has larger values of mean $NMI$. This indicates that the attributes noise have small impact on the ability of $BCMAG$ to detect the correct clusters when the graph modularity is high (i.e. structure mixing is low). In the same way, the bottom rows of the $BCMAG$ subplot indicate that when there is no attributes noise (or the noise level is small), $BCMAG$ is able to work well also for very small modularities, performing better than all the other methods.
We can conclude that our proposed strategy $BCMAG$ outperforms $BCMG$ and all the other methods when employing vertices attributes with no noise, improving the accuracy in recovering the right clusters also for very large mixing parameter. On the contrary, the impact of the attributes noise generally increases for small values of mixing parameter.
Note that the plots in~\cite[Section 5.2, Figure 2 (e)]{EA2013} show that the {\em SA-cluster} method~\cite{ZCY2009,CZY2011}, where the graphs attribute augmentation used in this paper was first introduced, has the worst performance on these benchmark graphs, not being able to recover the correct clusters. This demonstrates the benefits of our embedding strategy and of the new vector-valued distance as a valid alternative to distance-based approach relying on random walks on attribute-augmented graphs.

\subsection{Real networks}
\label{sub:realgraphs}

In this section we show the application of our new method on graphs originating by two real data networks that we indicate in the following as $Lazega \ Lawyers $ and $Yeast$.
A few indicators for these two networks are provided in Table \ref{table:RealNet}. In particular we indicate the network names, number of vertices $|V|$, number of edges $|E|$, sparsity (defined as $\frac{|E|}{|V|(|V|-1)/2}$), average vertex degree $c$, and number of attributes. Further details on experimental setting and discussion of results follow in the subsections.

\begin{table}[h]
\begin{center}
\begin{tabular}{||l| l l l l l||}
 \hline\hline
 Name & $|V|$ & $|E|$ & sparsity & $c$ & $m$ \\
 \hline\hline
 Yeast    & 2375 & 11693 & 0.004 &	9.85 & 1 \\
 \hline
 adv    & 71   &717    &0.289    &20.20 & 7 \\
 \hline
 adv36  &36    &289    &0.459  &16.06  & 7  \\
 \hline
 friend36 &36    &187    &0.297  &10.39  & 7 \\
 \hline\hline
\end{tabular}
\caption{Features of real networks. For each network we show the name, number of vertices  $|V|$, number of edges $|E|$, sparsity, average degree $c$ and number of available attributes $m$ }\label{table:RealNet}
\end{center}
\end{table}

\subsubsection*{Lazega Lawyers dataset}

The data $Lazega \ Lawyers$ comes from a network study of corporate law partnership that was carried out in a Northeastern US corporate law firm between 1988-1991. It includes measurements of networks among the 71 attorneys (partners and associates) of the firm. In particular, we looked at the advice network (adv) and at the friendship network (friend). Various members' attributes are also part of the dataset, including seniority, formal status, office in which they work, gender, law school attended, individual performance measurements (hours worked, fees brought in), attitudes concerning various management policy options, etc.
The ethnography, organizational and network analyses of this case are available in~\cite{Lazega2001}. The number $36$ indicates that subset of data for $36$ partners only were considered. In our analysis we considered 7 attributes, namely: status (1=partner; 2=associate), gender (1=man; 2=woman), office (1=Boston; 2=Hartford; 3=Providence), firm years, age, practice (1=litigation; 2=corporate), law school (1: Harvard, Yale; 2: Ucon; 3: other). The 7 attributes have a different number of distinct values, as summarized in Table~\ref{table:LazegaLab}.
\begin{table}[h]
\begin{center}
\begin{tabular}{||l| c||}
 \hline\hline
 Attribute & Distinct values \\
 \hline\hline
 Status  & 2  \\
 \hline
 Gender  & 2  \\
 \hline
 Office  & 3   \\
 \hline
 Firm years  & 27   \\
 \hline
 Age & 33 \\
 \hline
 Practice & 2   \\
 \hline
 Law school & 3  \\
 \hline\hline
\end{tabular}
\caption{Attributes used in the analysis of the $Lazega \ Lawyers$ networks and the corresponding number of distinct values. }\label{table:LazegaLab}
\end{center}
\end{table}

\begin{figure*}
        \centering
        \begin{subfigure}[b]{0.485\textwidth}
            \centering
            \includegraphics[page=1, width=\textwidth]{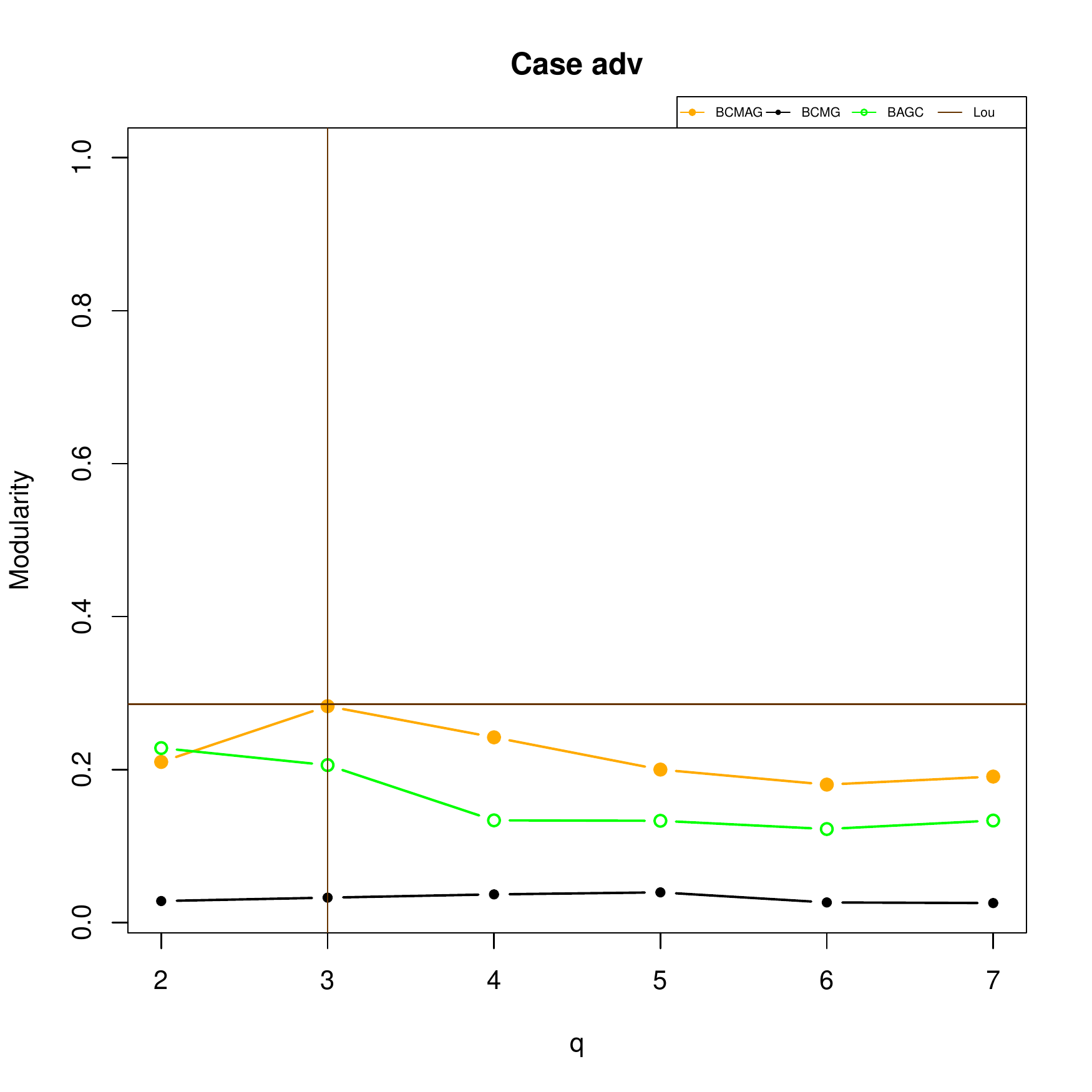}
            \caption[]%
            {{\small}}
            \label{fig:advQ}
        \end{subfigure}
        \hfill
        \begin{subfigure}[b]{0.485\textwidth}
            \centering
            \includegraphics[page=1, width=\textwidth]{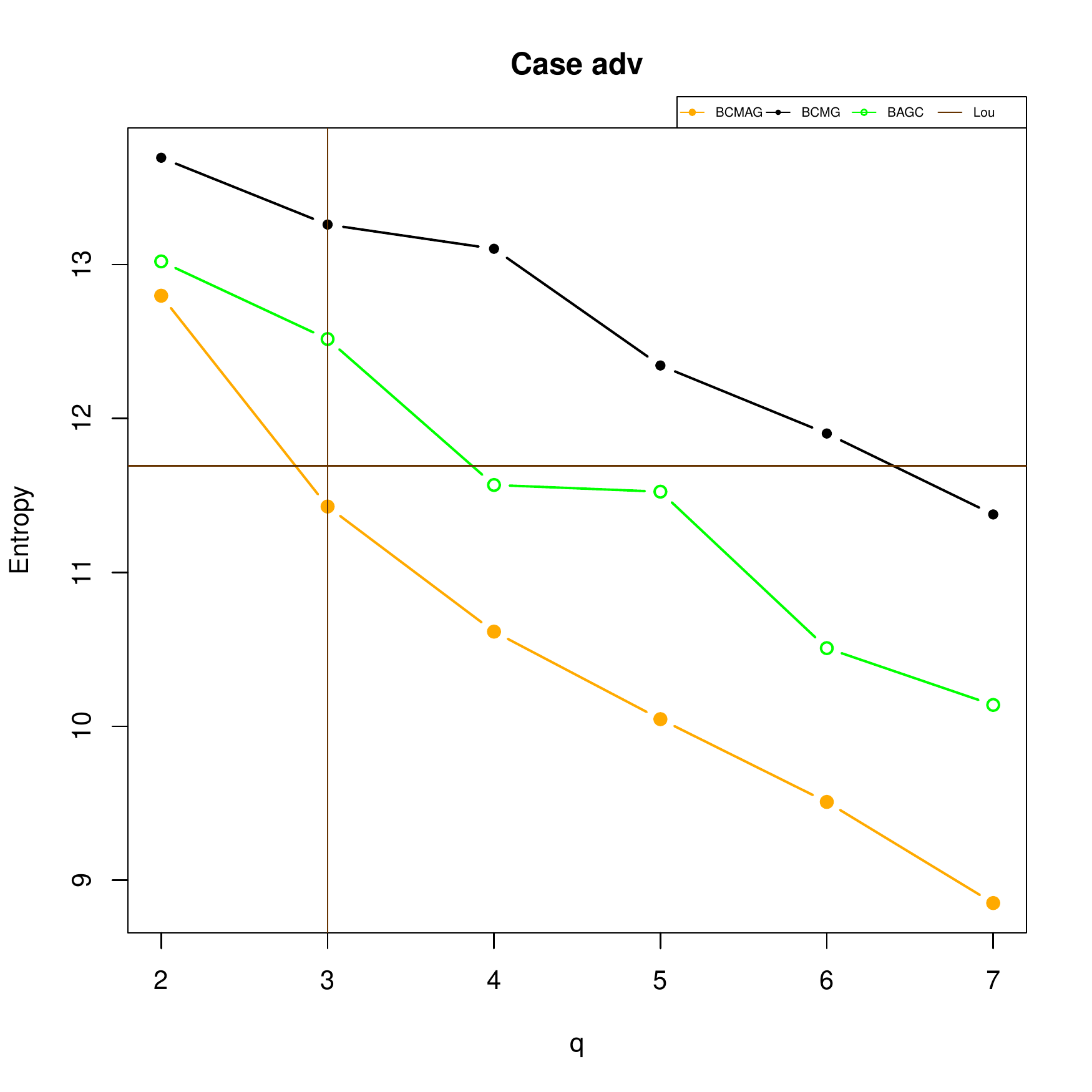}
            \caption[]%
            {{\small }}
            \label{fig:advE}
        \end{subfigure}

        \caption[]
       {\small Lazega layers dataset adv. Modularity (a) and entropy (b).}
        \label{fig:adv}
    \end{figure*}

\begin{figure*}
        \centering
        \begin{subfigure}[b]{0.485\textwidth}
            \centering
            \includegraphics[page=1, width=\textwidth]{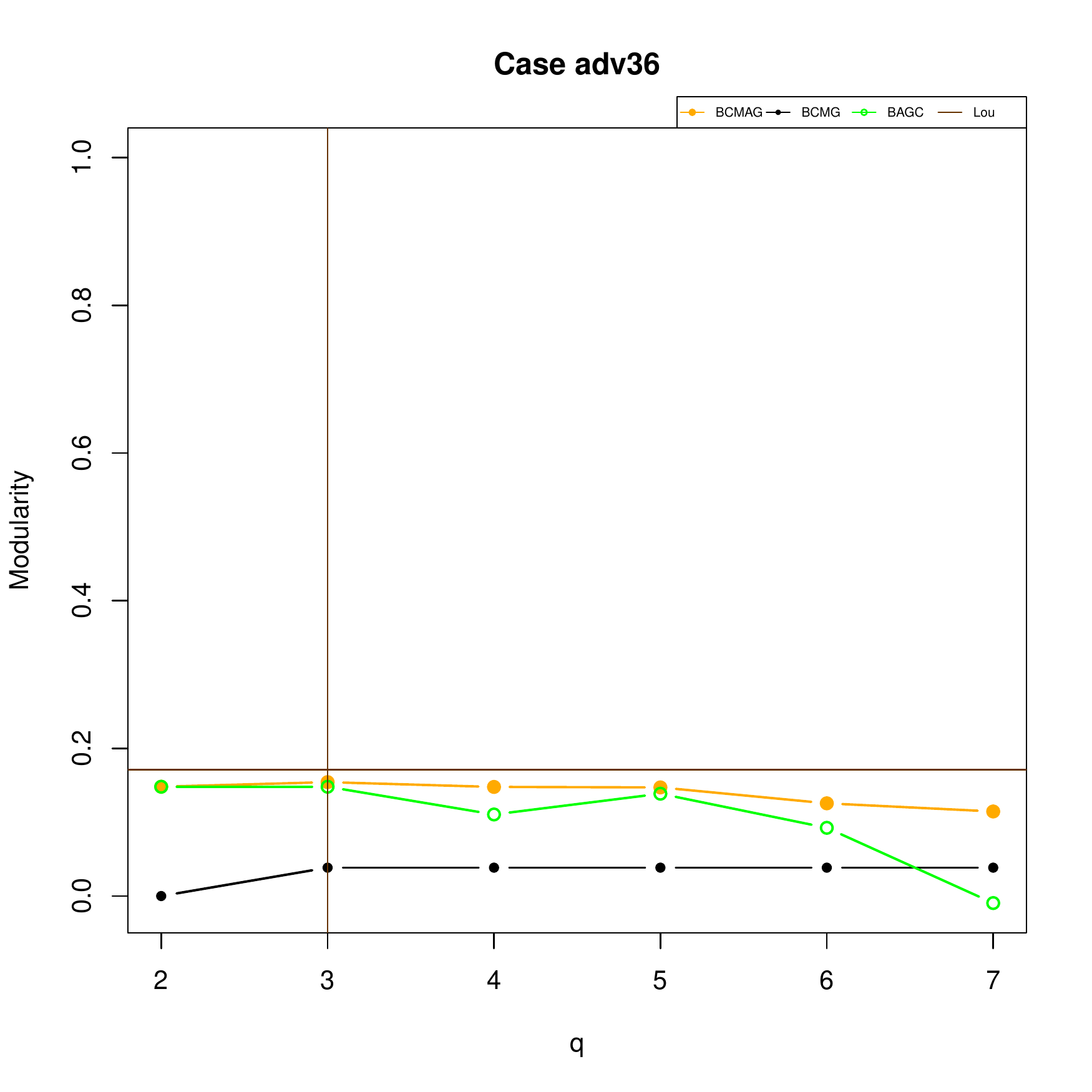}
            \caption[]%
            {{\small}}
            \label{fig:adv36Q}
        \end{subfigure}
        \hfill
        \begin{subfigure}[b]{0.485\textwidth}
            \centering
            \includegraphics[page=1, width=\textwidth]{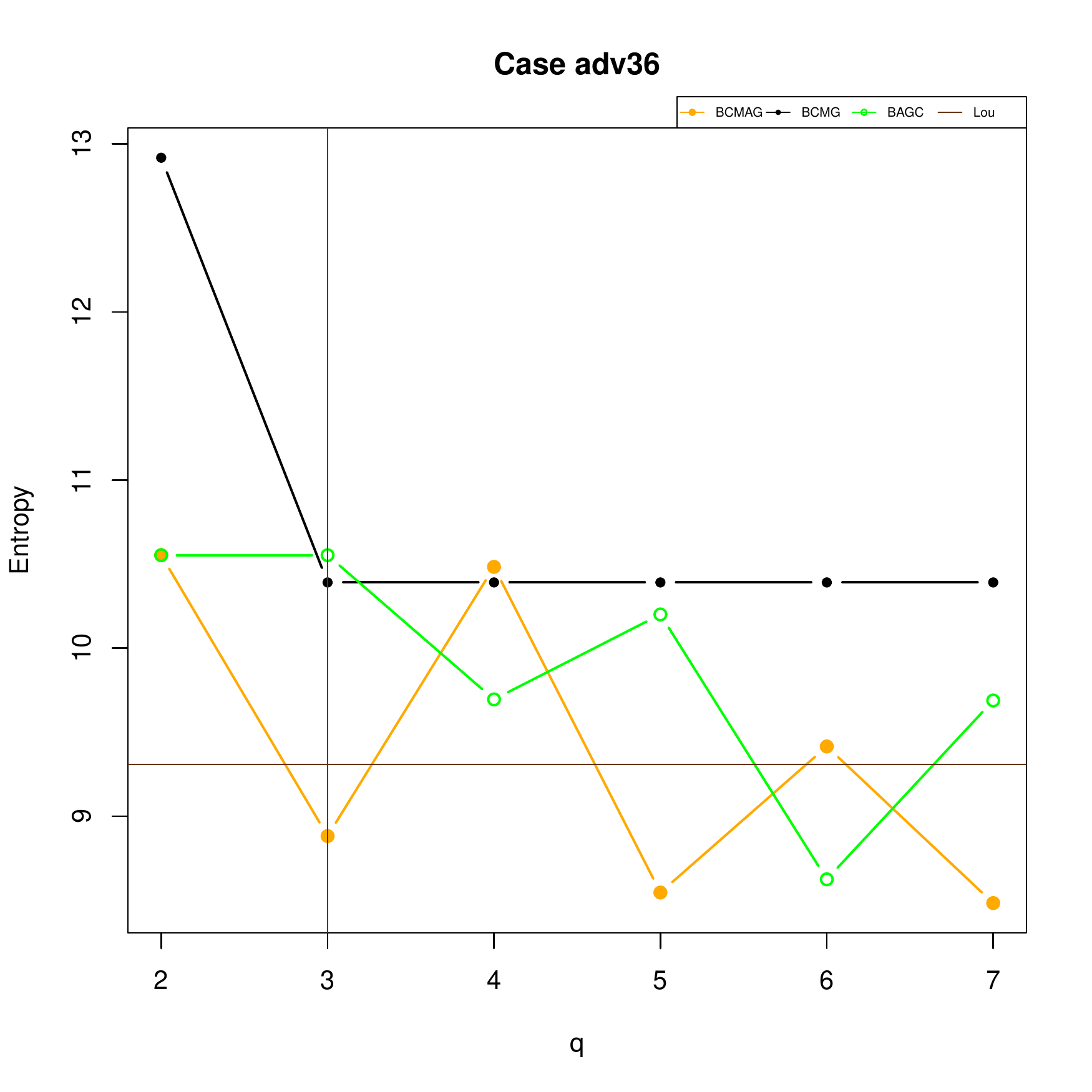}
            \caption[]%
            {{\small }}
            \label{fig:adv36E}
        \end{subfigure}

        \caption[]
       {\small  Lazega layers dataset adv36. Modularity (a) and entropy (b).}
        \label{fig:adv36}
    \end{figure*}

\begin{figure*}
        \centering
        \begin{subfigure}[b]{0.485\textwidth}
            \centering
            \includegraphics[page=1, width=\textwidth]{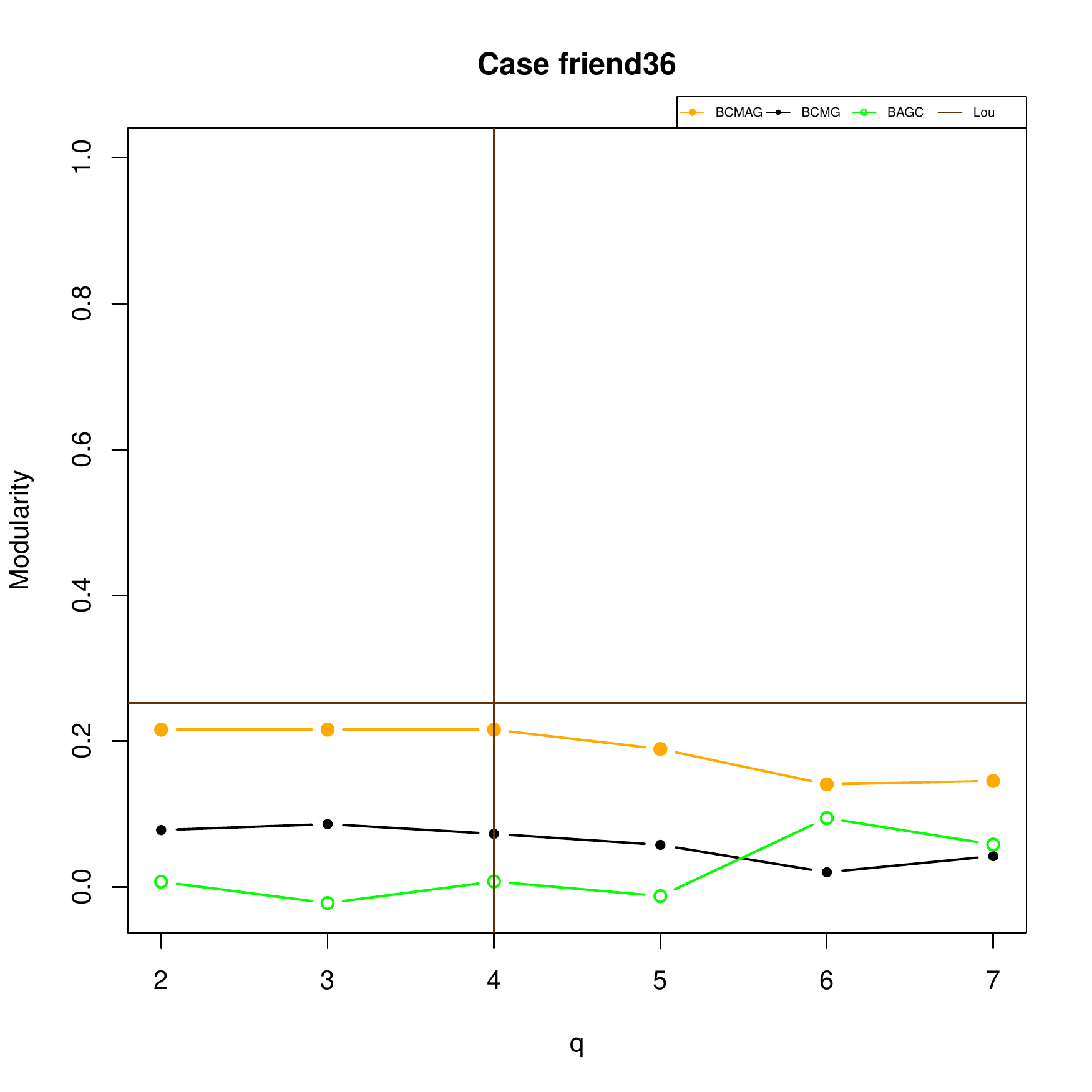}
            \caption[]%
            {{\small}}
            \label{fig:fr36Q}
        \end{subfigure}
        \hfill
        \begin{subfigure}[b]{0.485\textwidth}
            \centering
            \includegraphics[page=1, width=\textwidth]{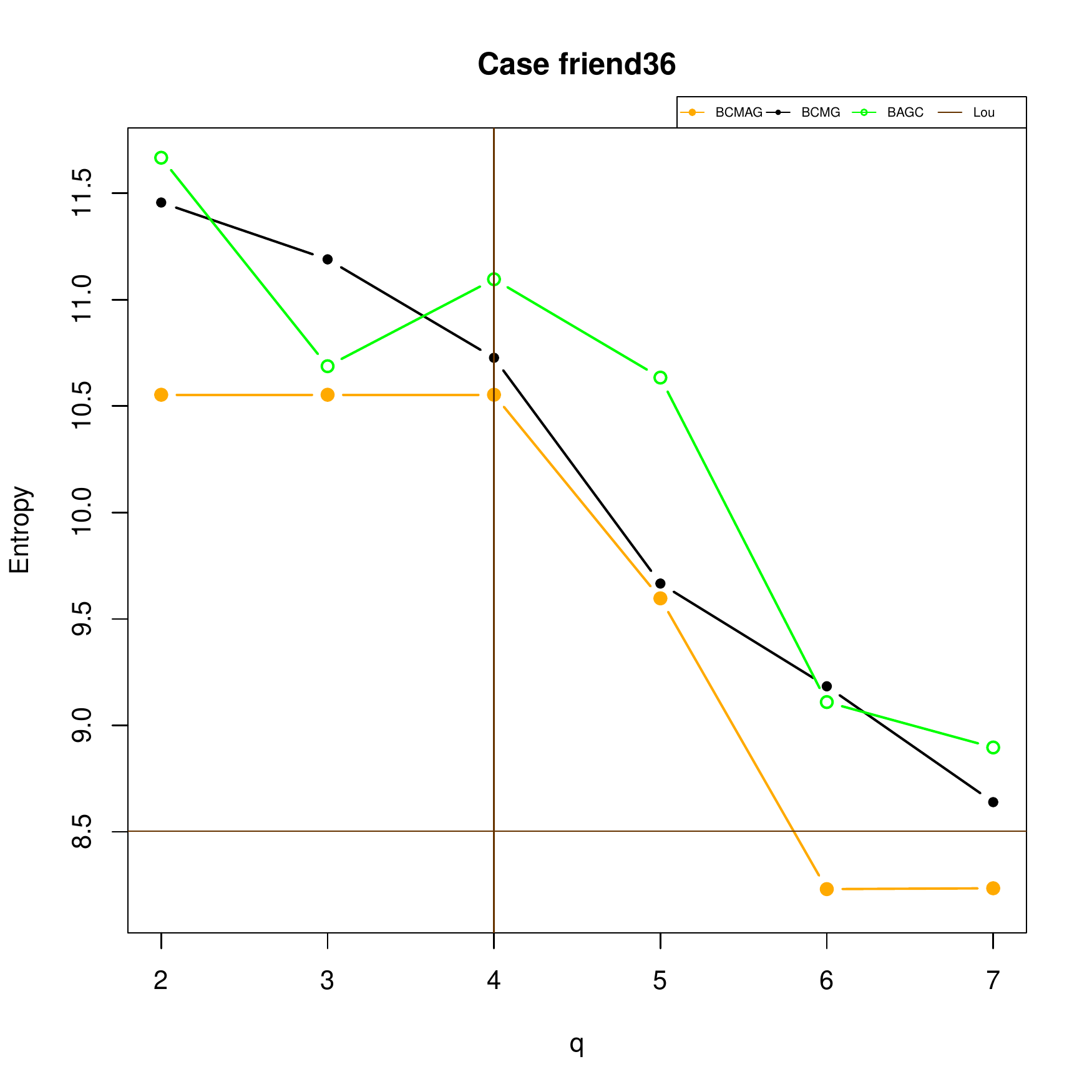}
            \caption[]%
            {{\small }}
            \label{fig:fr36E}
        \end{subfigure}

        \caption[]
       {\small  Lazega layers dataset friend36. Modularity (a) and entropy (b).}
        \label{fig:fr36}
    \end{figure*}

In Figures~\ref{fig:adv}-\ref{fig:fr36}, we plot the modularity and the entropy associated to the methods $BCMAG$, $BCMG$, $BAGC$ and $Lou$, respectively in subfigure (A) and (B) for different choice of the number $q$ of clusters. The $7$ categories of attributes considered in the analysis of the 3 $Lazega \ Lawyers$ networks show a different number of distinct values, and hence there is not an obvious choice of the number of clusters based on the attributes. As a consequence we show the number of clusters suggested by the $Lou$ method and the associated modularity and entropy, computed as in~(\ref{entr}). As we can see, the modularity of $BCMAG$ is on average the closest to $Lou$ clustering in every case. On the other hand, the entropy of $BCMAG$ is the lowest for all cases but the adv36, where oscillations are observed when $4$ and $6$ clusters are built. In all cases the new method appears as a large improvement over our previous method $BCMG$ which does not employ attributes information. It is worth noticing, that in all cases $BCMAG$ shows the best entropy value when $q=7$, indicating a good agreement with the partition based on attribute categories.

\subsubsection*{Yeast dataset}

\begin{figure*}
        \centering
        \begin{subfigure}[b]{0.485\textwidth}
            \centering
            \includegraphics[page=1, width=\textwidth]{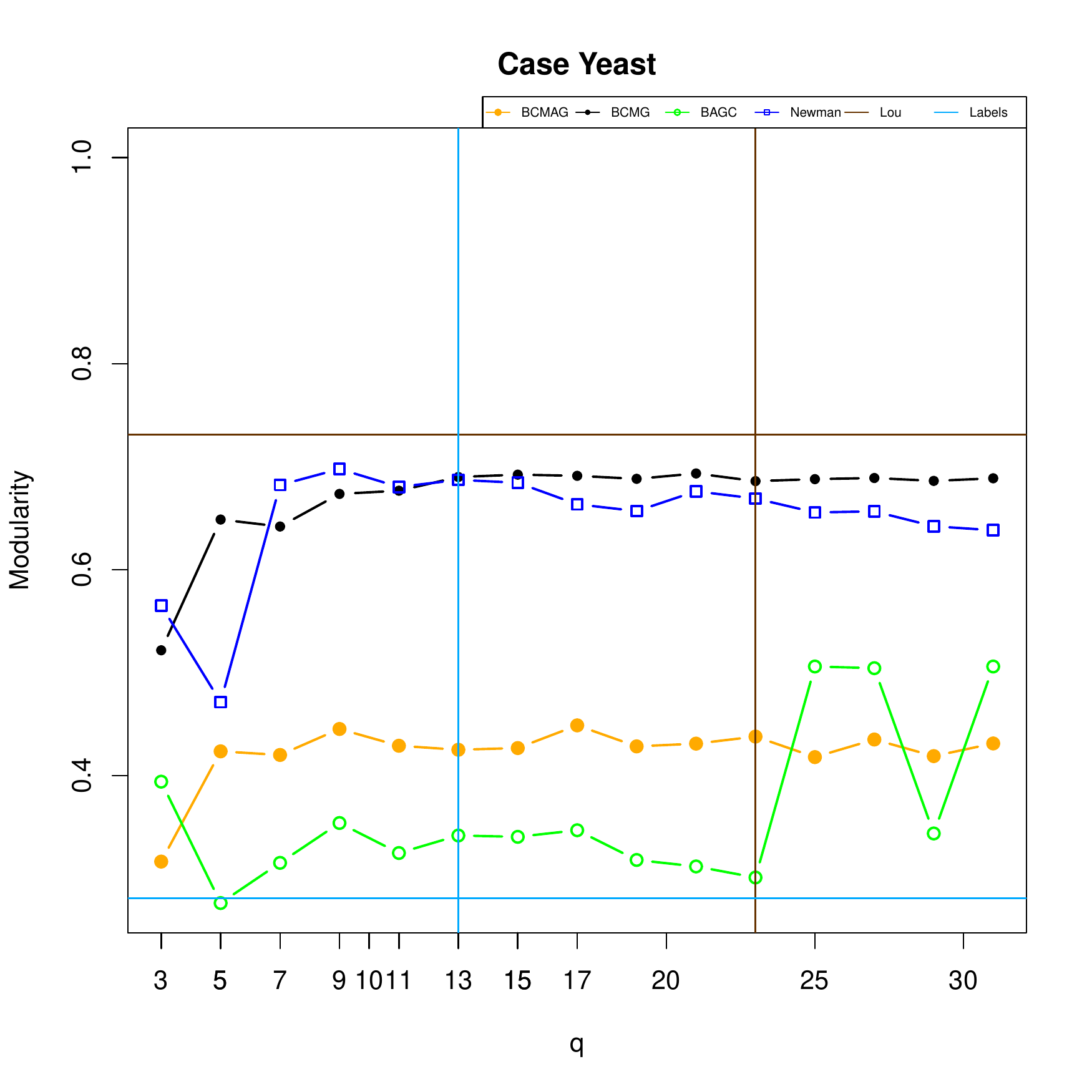}
            \caption[]%
            {{\small}}
            \label{fig:yeastQ}
        \end{subfigure}
        \hfill
        \begin{subfigure}[b]{0.485\textwidth}
            \centering
            \includegraphics[page=1, width=\textwidth]{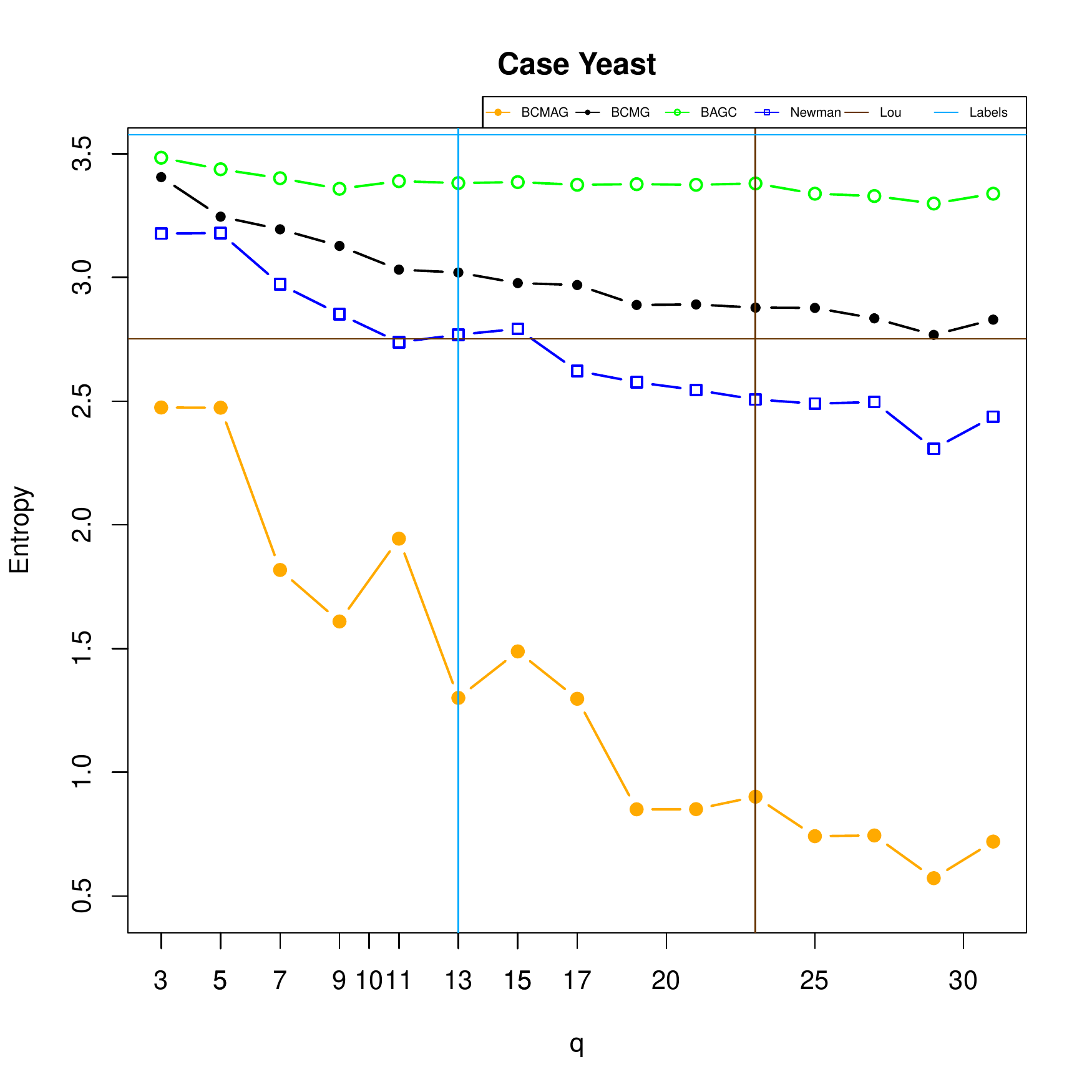}
            \caption[]%
            {{\small }}
            \label{fig:yeastE}
        \end{subfigure}

        \caption[]
       {\small Yeast dataset. Modularity (a) and entropy (b).}
        \label{fig:yeast}
    \end{figure*}

In this subsection, we apply the described methodologies to a comprehensive protein-protein interaction network, that we refer to as $Yeast$. This biological network aims to reveal many aspects of the complex regulatory network underlying the cellular function.
This data set was compiled by~\cite{Mering2002}, combining various sources. Only the interactions that have ‘‘high'' and ‘‘medium'' confidence are included in the analysis.
The data were downloaded from~\cite{yeastdata} and are also available in the R package $igraphdata$. We analysed the maximum connected component using as metadata the single label of the $13$ protein classes described in~\cite{yeastattributes}. Note that $39$ out of the $2375$ network proteins had not given class label. We considered each of them as belonging to an individual class. This resulted in a total of $2375+13+39$ augmented graph nodes.
In Figure~\ref{fig:yeast} we plot the modularity (sub-figure (a)) and the entropy (sub-figure (b)) corresponding to the clustering provided by $BCMAG$, $BCMG$, $BAGC$, $Newman$, computed as in~(\ref{condentr}) when $\C'$ is the ground truth represented by the attribute labels. We also provide values obtained by $Lou$ and the attribute labels (as the ground truth), in this case the entropy is computed as in~(\ref{entr}).
In this example, as said, we have one single set of labels indicating the protein class. Intuitively, we could consider as number of clusters $q=13$ (vertical cyan line), corresponding to the distinct classes provided in the labels.
Alternatively, as in the previous example, we could use the number of clusters $q=23$ found by $Lou$ (vertical brown line). As a consequence the most interesting area of the plots falls in the area $13\le q \le 23$. As we can see in sub-figure (a), the lowest modularity corresponds to the protein labels clustering. This indicates that the attributes provide low separation between the proteins in this case. Note that the modularity of all the methodologies is less oscillating when $13\le q \le 23$. Moreover, the modularity values provided by $Newman$ and $BCMG$ are closer to the modularity of $Lou$, while $BCMAG$ and $BAGC$ provide modularity values that are closer to the labels modularity. In particular, the $BCMAG$ modularity curve offers a trade off between using the labels or not. Indeed the $BCMAG$ modularity values are on average halfway between the horizontal cyan line ($Lou$ modularity) and the brown horizontal line (labels modularity).

If we look at the entropy in sub-figure (b), the general trend of the 4 methodologies $BCMAG$, $BCMG$, $BAGC$ and $Newman$ is to have a decreasing entropy (conditioned on the true labels) for increasing number of clusters $q$. The lowest entropy is reached by $BCMAG$ for any $q\in [3,31]$. The maximum absolute entropy is associated with the labels partition (cyan horizontal line). As a consequence $BCMAG$ has the maximum gain and hence appears as the most informative clustering.

\section{Conclusions}
\label{conc}

In this work we extend a clustering method for undirected graphs to employ possible additional information, such as vertex attributes, in order to improve clustering detection and quality.
The new method relies on an already proposed augmentation of the graph which incorporates vertex attributes in the graph in terms of new vertices and new edges. Starting from this augmentation we embed the graph in a low-dimensional vector space spanned by some near-null space vectors computed by an adaptive AMG process approximating small eigenvalues of the augmented graph Laplacian. Each such vector is appropriately used to build $1+m$ coordinates, where $m$ is the number of attributes, of the original graph vertices, so that vertices sharing an attribute value are in the same hyperplane defined by that value.
We also defined a modified Euclidean distance so that the new method was able to detect clusters by applying a modified K-means optimization algorithm. Experimental results show that the new method largely outperforms the previous approach which ignores the attributes for low-modularity graphs and it is also better than some of the available methods which fuse attribute and structure information, also including the {\em SA-cluster} which is considered the most representative of the {\em early-fusion} methods. One limitation of our approach can rely on the fact that it can only be computationally feasible for categorical attributed networks. Moreover, our approach would require large memory footprint and computational cost for very large networks, when the number of distinct attributes values is high. Feature selection or attributes transformation could be explored to address these issues. Further development would be needed to handle missing data or sparse attributes.

\section{Acknowledgments}

%  \begin{acks}
  The authors thank Clara De Santis, post-graduate fellow at IAC-CNR for helping in implementing some of the functions dealing with attributed graphs and included in the code \verb|BCMatch4Graphs|. We also thank Kamal Berahmand and Haithum Elhadi for sharing with us their codes implementing the LFR-EA model to generate graph samples.
 % \end{acks}

%%
%% The next two lines define the bibliography style to be used, and
%% the bibliography file.
\bibliographystyle{elsarticle-num}
\bibliography{bibattnet}

\begin{thebibliography}{10}
\expandafter\ifx\csname url\endcsname\relax
  \def\url#1{\texttt{#1}}\fi
\expandafter\ifx\csname urlprefix\endcsname\relax\def\urlprefix{URL }\fi
\expandafter\ifx\csname href\endcsname\relax
  \def\href#1#2{#2} \def\path#1{#1}\fi

\bibitem{CZY2011}
H.~Cheng, Y.~Zhou, J.~Yu, Clustering large attributed graphs: A balance between
  structural and attribute similarities, {ACM} Trans. Knowl. Discov. 2 (2011)
  12.1--12.33.
\newblock \href {https://doi.org/10.1145/1921632.1921638}
  {\path{doi:10.1145/1921632.1921638}}.

\bibitem{ZCY2009}
Y.~Zhou, H.~Cheng, J.~Yu, Graph clustering based on structural/attribute
  similarities, in: Proc. VLDB Endowment, Vol. 2 (1) of VLDB ‘09, August
  24-28, 2009, VLDB Endowment, Lyon, France, 2009, pp. 718--729.
\newblock \href {https://doi.org/10.14778/1687627.1687709}
  {\path{doi:10.14778/1687627.1687709}}.

\bibitem{S2007}
S.~Schaeffer, Graph clustering, Computer Science Review 1 (2007) 27--64.
\newblock \href {https://doi.org/10.1016/j.cosrev.2007.05.001}
  {\path{doi:10.1016/j.cosrev.2007.05.001}}.

\bibitem{F2010}
S.~Fortunato, Community detection in graphs, Physics Reports 486 (2010)
  75--174.
\newblock \href {https://doi.org/10.1016/j.physrep.2009.11.002}
  {\path{doi:10.1016/j.physrep.2009.11.002}}.

\bibitem{NC2011}
M.~Nascimento, A.~de~Carvalho, Spectral methods for graph clustering - a
  survey, European Journal of Operational Research 211 (2011) 221--231.
\newblock \href {https://doi.org/10.1016/j.ejor.2010.08.012}
  {\path{doi:10.1016/j.ejor.2010.08.012}}.

\bibitem{DCV2019}
P.~D'Ambra, L.~Cutillo, P.~Vassilevski, Bootstrap {AMG} for spectral
  clustering, Computational and Mathematical Methods 1 (2019) e1020.
\newblock \href {https://doi.org/10.1002/cmm4.1020}
  {\path{doi:10.1002/cmm4.1020}}.

\bibitem{HTF2001}
T.~Hastie, R.~Tibshirani, J.~Friedman, The element of statistical learning,
  Springer, New York, USA, 2001.

\bibitem{BCMM2015}
C.~Bothorel, J.~D. Cruz, M.~Magnani, B.~Micenkov\'a., Clustering attributed
  graphs: models, measures and methods, Network Science 3 (3) (2015) 408--444.
\newblock \href {https://doi.org/10.1017/nws.2015.9}
  {\path{doi:10.1017/nws.2015.9}}.

\bibitem{C2020}
P.~Chunaev, Community detection in node-attributed social networks: A survey,
  Computer Science Review 37, 100286 (2020) 1--24.
\newblock \href {https://doi.org/10.1016/j.cosrev.2020.100286}
  {\path{doi:10.1016/j.cosrev.2020.100286}}.

\bibitem{CGB2020}
P.~Chunaev, T.~Gradov, B.~K., Community detection in node-attributed social
  networks: how structure-attributes correlation affects clustering quality,
  Procedia Computer Science 178 (2020) 355--364.
\newblock \href {https://doi.org/10.1016/j.procs.2020.11.037}
  {\path{doi:10.1016/j.procs.2020.11.037}}.

\bibitem{BAGC2012}
Z.~Xu, Y.~Ke, Y.~Wang, H.~Cheng, J.~Cheng, A model-based approach to attributed
  graph clustering, in: Proceedings of the 2012 ACM SIGMOD International
  Conference on Management of Data, SIGMOD '12, Association for Computing
  Machinery, New York, NY, USA, 2012, p. 505–516.
\newblock \href {https://doi.org/10.1145/2213836.2213894}
  {\path{doi:10.1145/2213836.2213894}}.

\bibitem{BAGC2014}
Z.~Xu, Y.~Ke, Y.~Wang, H.~Cheng, J.~Cheng, Gbagc: A general bayesian framework
  for attributed graph clustering, ACM Trans. Knowl. Discov. Data 9, 5 (2014)
  1--43.
\newblock \href {https://doi.org/10.1145/2629616} {\path{doi:10.1145/2629616}}.

\bibitem{HuangYLLC17}
Z.~Huang, Y.~Ye, X.~Li, F.~Liu, H.~Chen,
  \href{https://doi.org/10.1007/978-3-319-57454-7\_29}{Joint weighted
  nonnegative matrix factorization for mining attributed graphs}, in: J.~Kim,
  K.~Shim, L.~Cao, J.~Lee, X.~Lin, Y.~Moon (Eds.), Advances in Knowledge
  Discovery and Data Mining - 21st Pacific-Asia Conference, {PAKDD} 2017, Jeju,
  South Korea, May 23-26, 2017, Proceedings, Part {I}, Vol. 10234 of Lecture
  Notes in Computer Science, 2017, pp. 368--380.
\newblock \href {https://doi.org/10.1007/978-3-319-57454-7\_29}
  {\path{doi:10.1007/978-3-319-57454-7\_29}}.
\newline\urlprefix\url{https://doi.org/10.1007/978-3-319-57454-7\_29}

\bibitem{Neville03}
J.~Neville, M.~Adler, D.~Jensen, Clustering relational data using attribute and
  link information, in: In Proceedings of the Text Mining and Link Analysis
  Workshop, 18th International Joint Conference on Artificial Intelligence,
  2003, pp. 9--15.

\bibitem{Combe2012}
D.~Combe, C.~Largeron, E.~Egyed-Zsigmond, M.~G\'{e}ry, {Combining relations and
  text in scientific network clustering}, in: First International Workshop on
  Semantic Social Network Analyis and Design at IEEE/ACM International
  Conference on Advances in Social Networks Analysis and Mining, 2012, pp.
  1280--1285.

\bibitem{NC2016}
M.~Newman, A.~Clauset, Structure and inference in annotated networks, Nature
  Communications 7:11863 (2016).
\newblock \href {https://doi.org/10.1038/ncomms11863}
  {\path{doi:10.1038/ncomms11863}}.

\bibitem{vL2007}
U.~von Luxburg, A tutorial on spectral clustering, Statistics and Computing 17
  (2007) 395--416.
\newblock \href {https://doi.org/10.1007/s11222-007-9033-z}
  {\path{doi:10.1007/s11222-007-9033-z}}.

\bibitem{PSZ2015}
R.~Peng, H.~Sun, P.~Zanetti, Partitioning well-clustered graphs: spectral
  clustering works!, Journal of Machine Learning Research: Workshop and
  Conference Proceedings 40 (2015) 1--33.

\bibitem{MLBFP2008}
P.~S. Vassilevski, Multilevel block factorization preconditioners, matrix-based
  analysis and algorithms for solving finite element equations, Springer, New
  York, USA, 2008.

\bibitem{DV2013}
P.~D'Ambra, P.~Vassilevski, Adaptive {AMG} with coarsening based on compatible
  weighted matching., Computing and Visualization in Science 16 (2013) 59--76.
\newblock \href {https://doi.org/10.1007/s00791-014-0224-9}
  {\path{doi:10.1007/s00791-014-0224-9}}.

\bibitem{DFV2018}
P.~D'Ambra, S.~Filippone, P.~Vassilevski, Boot{CM}atch: a software package for
  bootstrap {AMG} based on graph weighted matching, {ACM} Trans. Math. Softw.
  44-39 (2018) 1--25.
\newblock \href {https://doi.org/10.1145/3190647} {\path{doi:10.1145/3190647}}.

\bibitem{DV2019}
P.~D'Ambra, P.~Vassilevski, Improving solve time of aggregation-based adaptive
  {AMG}, Numerical Linear Algebra with Applications 26 (2019) e2269.
\newblock \href {https://doi.org/10.1002/nla.2269}
  {\path{doi:10.1002/nla.2269}}.

\bibitem{KV2013}
V.~Kuhlemann, P.~Vassilevski, Improving the communication pattern in mat-vec
  operations for large scale-free graphs by disaggregation, {SIAM} Journal on
  Scientific Computing 35 (2013) S465--S486.
\newblock \href {https://doi.org/10.1137/12088313X}
  {\path{doi:10.1137/12088313X}}.

\bibitem{Blondel2008}
D.~V. Blondel, J.~L. Guillaume, R.~Lambiotte, E.~Lefebvre, Fast unfolding of
  communities in large networks, Journal of Statistical Mechanics Theory and
  Experiment 2008 (2008) P10008.
\newblock \href {https://doi.org/10.1088/1742-5468/2008/10/P10008}
  {\path{doi:10.1088/1742-5468/2008/10/P10008}}.

\bibitem{CNM2004}
A.~Clauset, M.~Newman, C.~Moore, Finding community structure in very large
  networks, Physical Review, E, 70 (2004) 066111.
\newblock \href {https://doi.org/10.1103/PhysRevE.70.066111}
  {\path{doi:10.1103/PhysRevE.70.066111}}.

\bibitem{bacgcode}
X.~Z., Bayesian attributed graph clustering ({BAGC} code),
  \url{https://github.com/zhiqiangxu2001/BAGC} (2012).

\bibitem{Dannon2005}
L.~Danon, J.~Duch, A.~Diaz-Guilera, A.~Arenas, Comparing community structure
  identification, Journal of Statistical Mechanics: Theory and Experiment 2005
  (2005) P09008.
\newblock \href {https://doi.org/10.1088/1742-5468/2005/09/P09008}
  {\path{doi:10.1088/1742-5468/2005/09/P09008}}.

\bibitem{holland1983}
P.~W. Holland, K.~Laskey, S.~Leinhardt, Stochastic blockmodels: First steps,
  Social Networks 5-2 (1983) 109--137.
\newblock \href {https://doi.org/10.1016/0378-8733(83)90021-7}
  {\path{doi:10.1016/0378-8733(83)90021-7}}.

\bibitem{Decelle2011}
A.~Decelle, F.~Krzakala, C.~Moor, L.~Zdeborov\'a, Asymptotic analysis of the
  stochastic block model for modular networks and its algorithmic applications,
  Physical Review, E 84 (2011) 066106, 10.1103/PhysRevE.84.066106.
\newblock \href {https://doi.org/10.1103/PhysRevE.84.066106}
  {\path{doi:10.1103/PhysRevE.84.066106}}.

\bibitem{Newman2006}
M.~Newman, Modularity and community structure in networks, Proceedings of the
  National Academy of Sciences 103 (23) (2006) 8577--8582.
\newblock \href {https://doi.org/10.1073/pnas.0601602103}
  {\path{doi:10.1073/pnas.0601602103}}.

\bibitem{EA2013}
H.~Elhadi, G.~Agam, Structure and attributes community detection: comparative
  analysis of composite, ensemble and selection methods, in: Proceedings of the
  7th Workshop on Social Network Mining and Analysis, SNAKDD '13, August 2013,
  ACM, Chicago, Illinois (USA), 2013, pp. 10:1--10:7.
\newblock \href {https://doi.org/10.1145/2501025.2501034}
  {\path{doi:10.1145/2501025.2501034}}.

\bibitem{LFR2008}
A.~Lancichinetti, S.~Fortunato, F.~Radicchi,
  \href{https://link.aps.org/doi/10.1103/PhysRevE.78.046110}{Benchmark graphs
  for testing community detection algorithms}, Phys. Rev. E 78 (2008) 046110.
\newblock \href {https://doi.org/10.1103/PhysRevE.78.046110}
  {\path{doi:10.1103/PhysRevE.78.046110}}.
\newline\urlprefix\url{https://link.aps.org/doi/10.1103/PhysRevE.78.046110}

\bibitem{PS2020}
C.~Pizzuti, A.~Socievole, Multiobjective optimization and local merge for
  clustering attributed graphs, IEEE Transactions on Cybernetics 50~(12) (2020)
  4997--5009.
\newblock \href {https://doi.org/10.1109/TCYB.2018.2889413}
  {\path{doi:10.1109/TCYB.2018.2889413}}.

\bibitem{BMFM2022}
K.~Berahmand, M.~Mohammadi, A.~Faroughi, R.~P. Mohammadiani, A novel method of
  spectral clustering in attributed networks by constructing parameter-free
  affinity matrix, Cluster Computing 25 (2022) 869–888.
\newblock \href {https://doi.org/10.1007/s10586-021-03430-0}
  {\path{doi:10.1007/s10586-021-03430-0}}.

\bibitem{Lazega2001}
E.~Lazega, The Collegial Phenomenon: The Social Mechanisms of Cooperation Among
  Peers in a Corporate Law Partnership, Oxford University Press, Oxford, UK,
  2001.
\newblock \href {https://doi.org/10.1093/acprof:oso/9780199242726.001.0001}
  {\path{doi:10.1093/acprof:oso/9780199242726.001.0001}}.

\bibitem{Mering2002}
C.~von Mering, R.~Krause, B.~Snel, et~al., Comparative assessment of
  large-scale data sets of protein–protein interactions, Nature 417 (2002)
  399--403.
\newblock \href {https://doi.org/10.1038/nature750}
  {\path{doi:10.1038/nature750}}.

\bibitem{yeastdata}
Yeast data set,
  \url{http://www.nature.com/nature/journal/v417/n6887/suppinfo/nature750.html}.

\bibitem{yeastattributes}
Attributes for {Y}east data set,
  \url{https://rdrr.io/cran/igraphdata/man/yeast.html}.

\end{thebibliography}

\end{document}